\documentclass[10pt,journal,compsoc]{IEEEtran}
\usepackage{pifont}

\usepackage{ifpdf}

\ifCLASSOPTIONcompsoc
\usepackage[nocompress]{cite}
\else
\usepackage{cite}
\fi
\ifCLASSINFOpdf
\usepackage[pdftex]{graphicx}
\else
\usepackage[dvips]{graphicx}
\fi
\usepackage{amsmath}
\usepackage{algorithmic}

\usepackage{array}

\ifCLASSOPTIONcompsoc
\usepackage[caption=false,font=footnotesize,labelfont=sf,textfont=sf]{subfig}
\else
\usepackage[caption=false,font=footnotesize]{subfig}
\fi
\usepackage{url}

\usepackage{color}
\usepackage{lipsum}
\usepackage{algorithm}

\begin{document}
	\title{ResLT: Residual Learning for Long-tailed Recognition}
	\author{Jiequan~Cui,~\IEEEmembership{Student Member,~IEEE,}
		Shu~Liu,
		Zhuotao~Tian,~\IEEEmembership{Student Member,~IEEE,} \\
		Zhisheng~Zhong,~\IEEEmembership{Student Member,~IEEE,}
		Jiaya~Jia,~\IEEEmembership{Fellow,~IEEE} 
		
		\IEEEcompsocitemizethanks{\IEEEcompsocthanksitem J.~Cui, Z.~Tian, Z.~Zhong and J.~Jia are with the Department of Computer Science \& Engineering, The Chinese University of Hong Kong, ShaTin, Hong Kong.\protect\\
			E-mail: \{jqcui, zttian, zszhong21, leojia\}@cse.cuhk.edu.hk; \\ liushuhust@gmail.com 
			\IEEEcompsocthanksitem J.~Jia and S.~Liu are with the SmartMore.}}

	\IEEEtitleabstractindextext{%
		\begin{abstract}
			Deep learning algorithms face great challenges with long-tailed data distribution which, however, is quite a common case in real-world scenarios. Previous methods tackle the problem from either the aspect of input space (re-sampling classes with different frequencies) or loss space (re-weighting classes with different weights), suffering from heavy over-fitting to tail classes or hard optimization during training. To alleviate these issues, we propose a more fundamental perspective for long-tailed recognition, {\it i.e.}, from the aspect of parameter space, and aims to preserve specific capacity for classes with low frequencies. From this perspective, the trivial solution utilizes different branches for the head, medium, tail classes respectively, and then sums their outputs as the final results is not feasible. Instead, we design the effective residual fusion mechanism -- with one main branch optimized to recognize images from all classes, another two residual branches are gradually fused and optimized to enhance images from medium+tail classes and tail classes respectively. Then the branches are aggregated into final results by additive shortcuts. We test our method on several benchmarks, {\it i.e.}, long-tailed version of CIFAR-10, CIFAR-100, Places, ImageNet, and iNaturalist 2018. Experimental results manifest the effectiveness of our method. Our code is available at \url{https://github.com/jiequancui/ResLT}.
			
		\end{abstract}
		\begin{IEEEkeywords}
			Residual Learning, Imbalanced Learning, Long-tailed Recognition.
	\end{IEEEkeywords}}
	
	\maketitle
	\IEEEdisplaynontitleabstractindextext
	\IEEEpeerreviewmaketitle

	\IEEEraisesectionheading{\section{Introduction}\label{sec:introduction}}
	\IEEEPARstart{C}{onvolutional} neural networks (CNNs) have achieved impressive success on various tasks, including large-scale image classification \cite{vggnet, alexnet, googlenet, he2016deep, sandler2018mobilenetv2, ma2018shufflenet, hu2018squeeze, howard2017mobilenets, wang2018non, tan2019efficientnet}, object detection \cite{ren2015faster, liu2018path, lin2017feature, redmon2016you} and semantic segmentation \cite{Cordts_2016_CVPR, chen2018encoder, zhao2017pyramid}.  Especially, with the rise of neural network search \cite{DBLP:conf/cvpr/ZophVSL18, DBLP:conf/iclr/LiuSY19, DBLP:conf/cvpr/TanCPVSHL19, DBLP:conf/iccv/CuiCLLSJ19, DBLP:conf/iclr/CaiGWZH20, howard2019searching}, performance of CNNs have further taken a big step. However, the incredible progress stems in part from high-quality and large-scale datasets, such as ImageNet \cite{imagenet}, MS COCO \cite{coco} and Places \cite{zhou2017places}. These datasets are carefully designed with balanced distributions over different classes. In real-world applications, data could follow an unexpected long-tailed distribution where only a few head classes and a large number of tail classes exist. The long-tailed phenomenon may lead to severe degradation of performance for all models that do not take it into consideration. We in this paper address long-tailed recognition, {\it i.e.}, recognition on data with long-tailed distributions. 
	
	\begin{figure*}[t]
		\begin{center}
			\includegraphics[width=0.8\textwidth]{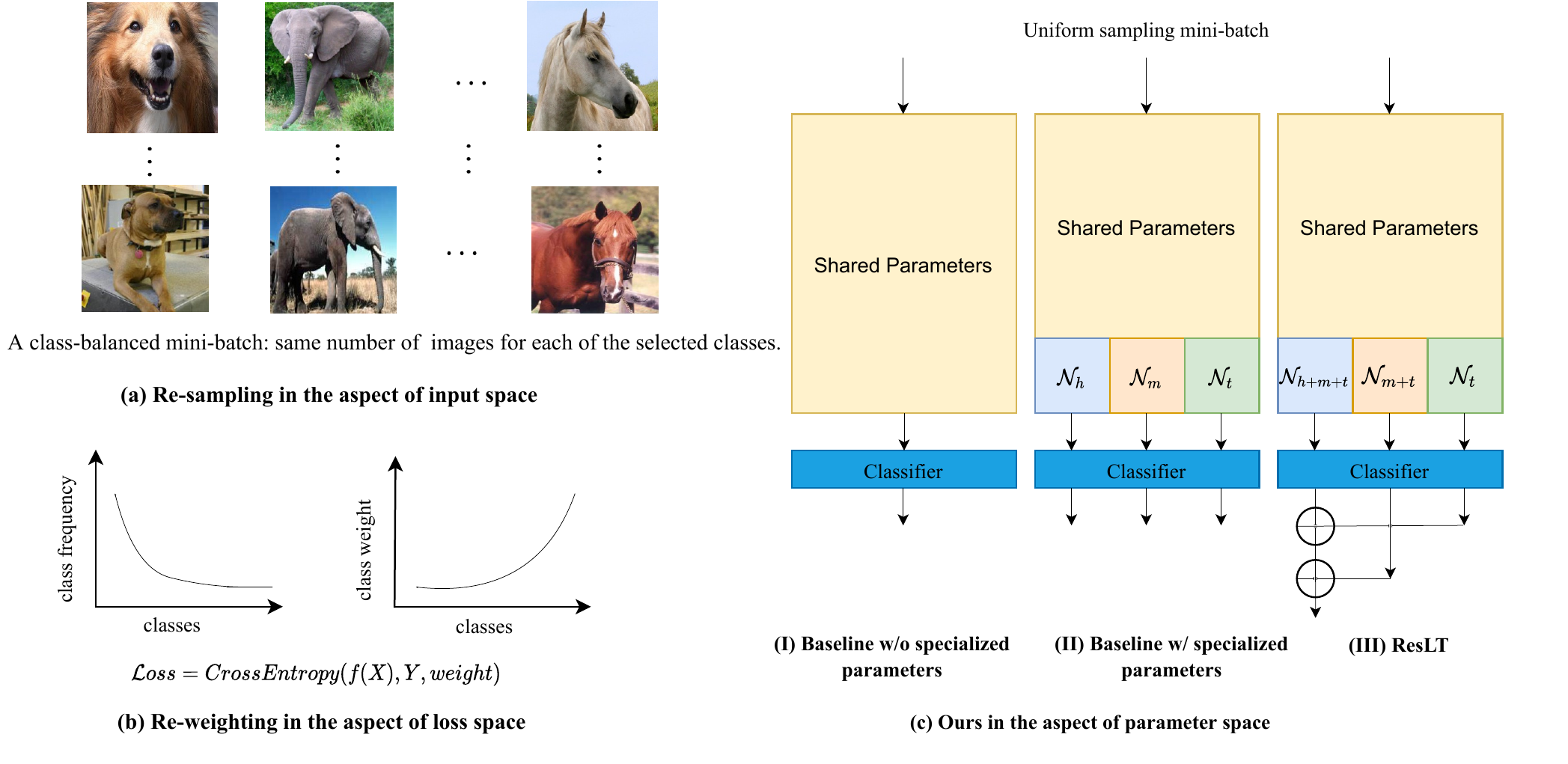}
			\caption{Comparison between our method and previous ones. Previous re-sampling and re-weighting balance input space and loss space respectively, with effect on the model parameters. Our method is built regarding the most fundamental parameter space, individually preserves specialized parameters for the head, medium, and tail classes with three sub-branches. These sub-branches are combined finally to enhance classification results of the tail and medium classes by the proposed residual fusion mechanism. In (b), $(X, Y)$ is a batch of images and their corresponding labels. $weight$ represents a vector of class-wise weights which usually is in reverse proportion to the number of samples of classes. The classifier is a fully connected (FC) layer that is shared among the three branches in (II) and (III) of (c).}
			\label{fig:comparison_methods}
		\end{center}
		\vspace{-0.1in}
	\end{figure*}
	
	One of the greatest challenges in the long-tailed setting is data imbalance. It is also the principal reason for head classes dominating the training procedure, making head classes enjoy much higher accuracy than tail classes. In the literature, two kinds of methods tackle the long-tailed problem via re-sampling and re-weighting \cite{DBLP:journals/nn/BudaMM18,DBLP:conf/cvpr/HuangLLT16,DBLP:conf/cvpr/CuiJLSB19, he2009learning,chawla2002smote, shen2016relay}. It is found that over-sampling tail-class images \cite{shen2016relay, zhong2016towards, buda2018systematic, byrd2019effect} may still suffer from heavy over-fitting to tail classes while under-sampling by dropping a large number of head-class images \cite{he2009learning, japkowicz2002class,buda2018systematic} inevitably impairs the generalization ability of deep models. Also, for re-weighting methods, previous works \cite{huang2016learning, huang2019deep} have demonstrated that re-weighting strategies may cause optimization difficulties during training on large-scale and real-world datasets.   
	
    Regarding procedures, re-sampling rebalances head and tail classes in input space by constructing balanced mini-batches during training as shown in Fig.~\ref{fig:comparison_methods}(a). Differently, re-weighting deals with the loss space by assigning different weights for classes according to the respective numbers of samples as shown in Fig.~\ref{fig:comparison_methods}(b). Albeit the procedural difference, we note that these two lines, by nature, both {\it eventually make effects on model parameters} to adjust tail classes response. This finding leads to our key idea of {\textit{re-balancing in parameter space directly.}}
	
	Our direct operations on the head and tail classes regarding parameter space can avoid heavy over-fitting to tail classes and the difficult optimization problem mentioned above. To illustrate that designing these effective operations is nontrivial, we show in Fig.~\ref{fig:comparison_methods}(c)(II) a naive solution. It preserves specialized parameters for the head, medium, and tail classes respectively via three branches of $\mathcal{N}_{h}$, $\mathcal{N}_{m}$ and $\mathcal{N}_{t}$. The branches are optimized for respective recognition and the final result is obtained by summing or averaging outputs of them. Unfortunately, this naive solution does not work well in experiments.
	
	Contrary to these naive operations, we model the long-tailed recognition problem as one residual learning process and propose a novel residual fusion mechanism, as shown in Fig.~\ref{fig:comparison_methods}(c)(III). It has one main branch ($\mathcal{N}_{h+m+t}$) optimized to classify images of all classes while the other two residual branches ($\mathcal{N}_{m+t}$ and $\mathcal{N}_{t}$) are optimized to recognize images in medium+tail and tail classes respectively. Outputs of these three branches are aggregated into the final result by additive shortcuts.
	
	An intuitive explanation of our strategy is to gradually enhance classification results on tail classes with other learned residual branches. It is noteworthy that samples belonging to the head, medium, and tail classes still dominate $\mathcal{N}_{h+m+t}$, $\mathcal{N}_{m+t}$ and $\mathcal{N}_{t}$ individually in the training procedure, and thus guarantee parameter specialization of the head, medium, and tail classes, which is coherent with the experimental phenomenon in \ref{sec:individual_branch_performance}.
	
	We extensively validate our method on several long-tailed benchmark datasets using long-tailed versions of CIFAR-10, CIFAR-100, ImageNet, Places, and iNaturalist 2018 data. Experimental results manifest the effectiveness of our method for long-tailed recognition.
	Our key contributions are as follows.
	\begin{itemize}
		\item We study the problem from a new perspective of parameter space that leads to an effective re-balance between head and tail classes in the long-tailed setting.
		\item We propose the residual fusion mechanism, making re-balance in the aspect of parameter space feasible and alleviating limitations in previous works.
		\item We validate our method on representative benchmarks. Note that this is the first time that a one-stage method consistently surpasses two-stage methods on long-tailed CIFAR-10, CIFAR-100, ImageNet, iNaturalist 2018 all these datasets.
	\end{itemize}

	\section{Related work}\label{sec:related work}
	\vspace{+0.1in}
	\noindent {\bf Re-sampling Strategy}
	There are two groups of re-sampling strategies: over-sampling the tail class images \cite{shen2016relay, zhong2016towards, buda2018systematic, byrd2019effect} and under-sampling the head class images \cite{he2009learning, japkowicz2002class,buda2018systematic}. Over-sampling is regularly useful on large datasets and often suffers from heavy over-fitting to tail classes especially on small datasets. For under-sampling, it discards a large portion of data, which inevitably causes degradation of the generalization ability of deep models. Moreover, the under-sampling strategy is not suitable when data is largely imbalanced.
	
	\vspace{+0.1in}
	\noindent {\bf Re-weighting Strategy}
	Re-weighting \cite{huang2016learning, huang2019deep,wang2017learning,ren2018learning,shu2019meta,jamal2020rethinking} is another prominent strategy. It assigns different weights for classes and even samples. The vanilla re-weighting method gives class weights in reverse proportion to the number of samples of classes. However, with large-scale data, re-weighting makes the deep models difficult to optimize during training \cite{huang2016learning, huang2019deep}. Cui et al. \cite{DBLP:conf/cvpr/CuiJLSB19} relieved the problem using "effective numbers" to calculate class weights. Tan et al. \cite{tan2020equalization} conducted re-weighting from the perspective of gradients by ignoring the suppressed gradients of tail classes.
	
	Another line of work is to adaptively re-weight each sample. Focal loss \cite{focalloss} assigned smaller weights for well-classified samples. Li et al. \cite{li2019gradient} down-weighted samples with very small or large gradients in response to the principle that samples with small gradients are usually well-classified and samples with large gradients are usually out of the distribution.
	
	\vspace{+0.1in}
	\noindent {\bf Two-stage Methods}
	Cao et al. \cite{cao2019learning} first observed that re-weighting and re-sampling are inferior to the vanilla empirical risk minimization (ERM) algorithm before annealing the learning rate. A two-stage deferred re-balancing optimization schedule was proposed, which trains using vanilla ERM with the LDAM \cite{cao2019learning} loss before annealing learning rate, and then deploys a re-weighted LDAM loss with a much smaller learning rate.  
	
    Recently, Kang et al. \cite{kang2019decoupling} and Zhou et al. \cite{zhou2019bbn} concluded that although class re-balance strategies matter when jointly training representation and classifier, uniform sampling gives more general representations. Based on this observation, Kang et al. \cite{kang2019decoupling} achieved state-of-the-art results on long-tailed ImageNet (ImageNet-LT) by decomposing representation and classifier learning, {\it i.e.}, first train the deep models with uniform sampling, then fine-tune the classifier with class-balanced sampling while keeping parameters of representation learning fixed. Similarly, Zhou et al. \cite{zhou2019bbn} integrated {\it mixup} into the proposed cumulative learning strategy with which they bridged the representation learning and classifier re-balancing and achieved state-of-the-art results on long-tailed CIFAR (CIFAR-LT) and iNaturalist 2018. 
	
	The two-stage design defies the end-to-end merit that we used to believe since the deep learning era. But why does the two-stage training outperform the end-to-end one largely in long-tailed classification? The work \cite{tang2020long} by Tang et al. analyzed the reason from the perspective of causal graph and concluded that the bad momentum causal effects played a vital role. Differently, we address the problem in terms of residual learning.
	
	\vspace{+0.1in}
	\noindent {\bf Meta Learning and Transfer Learning} MetaModelNet \cite{wang2017learning}, by Wang et al., learned a meta regression network from head classes and used it to construct the classifier for tail classes. 
	OLTR \cite{Liu_2019_CVPR} used memory banks to store mid- and high-level features from head to tail classes. Then they transferred knowledge from head classes to tail classes with a dynamic-meta embedding mechanism. Abdullah Jamal et al. analyzed imbalanced learning from a domain adaptation perspective in \cite{jamal2020rethinking} and proposed a meta learning framework for long-tailed recognition. Saurabh et al. \cite{sharma2020long} trained individual experts and transferred knowledge from these experts to the student model. However, the performance with meta learning or transfer learning methods still far from two-stage methods and the training procedure is usually much more complicated.

	\section{Method}
	\vspace{+0.1in}
	\noindent {\bf Motivation} 
	As illustrated in Fig.~\ref{fig:comparison_methods}(a), re-sampling strategies build class-balanced mini-batch data for training. However, over-sampling tail class images is easy to get into trouble of heavy over-fitting to tail classes while under-sampling head class images hurts the generalization ability of deep models due to discarding a large number of head-class images that are conducive in learning good representations. For re-weighting methods, previous works \cite{huang2016learning, huang2019deep} demonstrate that it makes deep models difficult to optimize. We, unlike these solutions, explore another way to re-balance regarding parameter space directly, avoiding above disadvantages and making further improvements.
	
	\begin{table}[htp]
		\small
		\centering
		\caption{Top-1 accuracy (\%) of baselines on CIFAR-10-LT with the imbalance factor 0.02. ResNet-32 is adopted.}
		\resizebox{.95\linewidth}{!}
		{
			\begin{tabular}{|c|c|c|c|c|}
				\hline
				Method &Many-shot & Medium-shot &Few-shot &All\\
				\hline
				\hline
				Baseline (CE) &93.60 &76.33 &68.55 &78.50\\
				\hline
				\hline
				Baseline (1) &90.70 &54.73 &41.48 &60.22 \\
				Baseline (2) &90.83 &55.63 &40.75 &60.24 \\
				Baseline (3) &86.73 &45.50 &55.33 &61.80 \\
				\hline
				\hline
				ResLT (Ours)     &85.83 &81.47 &81.43 &\textbf{83.46} \\
				\hline
			\end{tabular}
		}
		\label{tab:parameter_sepcialization_results}
	\end{table}
	
	\subsection{Re-balancing in Parameter Space}
	\label{sec:parameter_specialization_baselines}
	We are the first to propose re-balancing regarding parameter space. To understand the difficulty, we show a naive way by preserving several specialized parameters for the head, medium, tail classes separately by three branches ({\it i.e.}, $\mathcal{N}_{h}$, $\mathcal{N}_{m}$, and $\mathcal{N}_{t}$). They are shown in Fig.~\ref{fig:comparison_methods}(c)(II). After sorting regarding the number of images, all the classes are divided into 3 groups for the head, medium, tail classes, and we guarantee that the three groups have the same imbalance factor $\beta$. $\beta$= $\frac{N_{min}}{N_{max}}$ where $N_{max}$ and $N_{min}$ are the numbers of training samples for the most frequent class and the least frequent class in each group. Under this case, the core issue is the way to fuse the outputs of these branches into the final prediction because it is impossible to know whether one image belongs to head or medium or tail classes at inference time. Therefore, fusing the three branches is not trivial. Here we compare three straightforward fusion methods with the basic vanilla baseline using cross-entropy. They are set as described in Algorithm \ref{algorithm_baselines}.
	
	\begin{algorithm}
	\caption{Pseudocode of baselines implementation in a PyTorch-like style.}
	\label{algorithm_baselines}
	\begin{algorithmic}[1]
	\STATE \textbf{Input:} $x \in \mathcal{R}^{c \times h \times w}$ is an image feature map from the network backbone. Specifically, $c$ is the number of channels, $h$ and $w$ are spatial dimensions. $\mathcal{N}_{h}(\cdot)$, $\mathcal{N}_{m}(\cdot)$, and $\mathcal{N}_{t}(\cdot)$ are three branches for the head, medium, and tail classes respectively. $cls(\cdot)$ is the shared linear classifier among the branches with the weight $w \in \mathcal{R}^{k \times c}$. $k$ is the number of classes. $\sigma(\cdot)$ is the softmax activation function.
	\STATE $out_{h}$, $out_{m}$, $out_{t}$ = $cls(\mathcal{N}_{h}(x))$, $cls(\mathcal{N}_{m}(x))$, $cls(\mathcal{N}_{t}(x))$;
	\STATE \textbf{Baseline (1)}:
	\STATE $out_{a}$ = torch.stack([$out_{h}$, $out_{m}$, $out_{t}$], dim=1);
	\STATE $pred$ = torch.argmax($out_{a}$.max(1), dim=1).
	
	\STATE \textbf{Baseline (2)}:
	\STATE $out_{a}$ = $out_{h}$ + $out_{m}$ + $out_{t}$;
	\STATE $pred$ = torch.argmax($out_{a}$, dim=1).
	
	\STATE \textbf{Baseline (3)}:
	\STATE $out_{a}$ = $\sigma(out_{h})$ + $\sigma(out_{m})$ + $\sigma(out_{t})$;
	\STATE $pred$ = torch.argmax($out_{a}$, dim=1).
	\end{algorithmic}
    \end{algorithm}
	
	The experimental results are summarized in the Table~\ref{tab:parameter_sepcialization_results}. It shows that these straightforward fusion methods do not work satisfyingly and even yield inferior results compared to the vanilla baseline. We instead propose an effective residual fusion mechanism to handle this challenging issue. 
	
	\begin{figure*}[t]
		\begin{center}
			\includegraphics[width=0.99\textwidth]{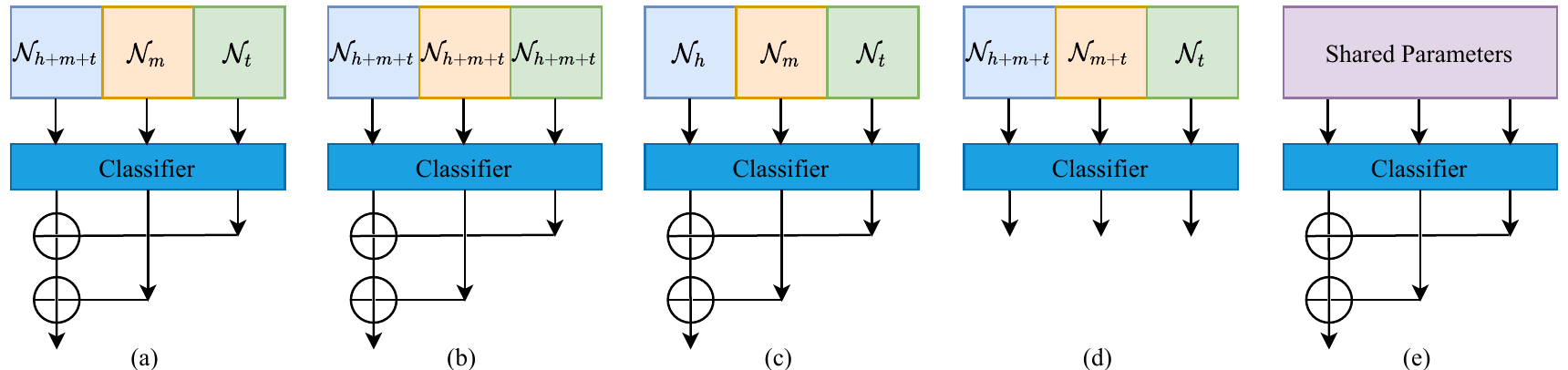}
			\caption{Ablation study for the residual fusion and parameter specialization mechanisms. (a), (b), (c), (d), and (e) are variants of ResLT. (a) is a weak version of ResLT. For (b) and (c), the three branches have no residual relationship -- nested class assignments, unlike ResLT. For (d), there is no additive shortcut, different from ResLT. (e) is without parameter specialization mechanism.}
			\label{fig:ablation_study_residual}
		\end{center}
		\vspace{-0.1in}
	\end{figure*}
	
	\subsection{Residual Fusion Mechanism}
	Under the long-tailed setting, models usually achieve high accuracy on head classes while the performance on tail classes is unsatisfactory. Based on this phenomenon, we model the long-tailed recognition problem as one residual learning process. It uses extra learned residual branches to enhance classification results on tail classes. Here we stress that the "residual" denotes the nested class assignments for different branches in our method.
	
	As shown in Fig.~\ref{fig:comparison_methods}(c)(III), the model parameters are divided into two parts: shared and specialized parameters. Shared parameters are for common features across all the classes while specialized parameters are to preserve specific capacity for the head, medium, tail classes with three branches $\mathcal{N}_{h+m+t}$, $\mathcal{N}_{m+t}$ and $\mathcal{N}_{t}$ correspondingly. In fact, the three branch networks are implemented by only one extra 1x1 grouped convolution with group number 3. Though only a few parameters are introduced, this parameter specialization mechanism plays an important role as demonstrated in Sec. \ref{sec:parameter_specialization} with experimental verification.  
	
	To effectively aggregate the three branches, we propose the residual fusion module shown in Fig.~\ref{fig:comparison_methods}(c)(III). We model the long-tailed recognition problem as such residual learning process where one main branch $\mathcal{N}_{h+m+t}$ learns to recognize images of all the classes and the other two residual branches $\mathcal{N}_{m+t}$, $\mathcal{N}_{t}$ are used to classify medium+tail classes and tail classes respectively. The outputs of the three branches are added together as the final result. It is worth noting that head, medium, and tail classes still dominate branches $\mathcal{N}_{h+m+t}$, $\mathcal{N}_{m+t}$, $\mathcal{N}_{t}$ respectively, and thus the parameter specialization of these classes are preserved. 
	
	In our residual fusion mechanism, loss functions are deployed in training process as
	\begin{small}
		\begin{align}
		&\mathcal{L}_{fusion} = \mathcal{J}(\mathcal{N}_{h+m+t}(X) \!+\! \mathcal{N}_{m+t}(X) \!+\! \mathcal{N}_{t}(X), Y), \label{eq:res_fusion_loss1}\\
		&\mathcal{L}_{branch} = \sum_{i \in \{h+m+t, m+t, t\}} \mathcal{J}(N_{i}(SX_{i}), SY_{i})		 
		\label{eq:res_fusion_loss2}, \\
		&\mathcal{L}_{all} = (1-\alpha) \mathcal{L}_{fusion} + \alpha \mathcal{L}_{branch},
		\label{eq:res_fusion_loss3}
		\end{align}
	\end{small} 
	where ($X$, $Y$) denotes the uniformly sampled images and labels of a mini-batch.  ($SX_{h+m+t}$, $SY_{h+m+t}$) is the same as ($X$, $Y$) consisting of all class images. ($SX_{m+t}$, $SY_{m+t}$) is a subset of $(X,Y)$ only containing images of medium and tail classes. ($SX_{t}$, $SY_{t}$) is a subset of $(X,Y)$ only containing images belonging to tail classes. $\mathcal{J}$ is cross-entropy loss and $\alpha$ is a hyper-parameter.
	
	The fusion loss item $\mathcal{L}_{fusion}$ in Eq. \eqref{eq:res_fusion_loss1} optimizes for all the classes. The outputs of branches $\mathcal{N}_{m+t}$ and $\mathcal{N}_{t}$ are added to the outputs of main branch $\mathcal{N}_{m+n+t}$ to obtain the fused outputs. Therefore, during inference, we just sum outputs of the three branches as the final result. 
	The branch-independent loss item $\mathcal{L}_{branch}$ in Eq. \eqref{eq:res_fusion_loss2} is for $\mathcal{N}_{h+m+t}$, $\mathcal{N}_{m+t}$ and $\mathcal{N}_{t}$ respectively, further encouraging parameter specialization for head, medium, and tail classes respectively. $\mathcal{L}_{all}$ is our final loss which is a weighted sum of $\mathcal{L}_{fusion}$ and $\mathcal{L}_{branch}$ with a trade-off hyper-parameter $\alpha$.
	
	\subsection{Analysis of Our Method}
	Our method depends on two key components: parameter specialization mechanism and residual learning mechanism. Here we analyze each of them with extensive experiments.
	
		\begin{figure*}[!t]
		\centering
		\subfloat[Importance of parameter specialization on CIFAR-10-LT.]{
			\includegraphics[width=0.23\textwidth]{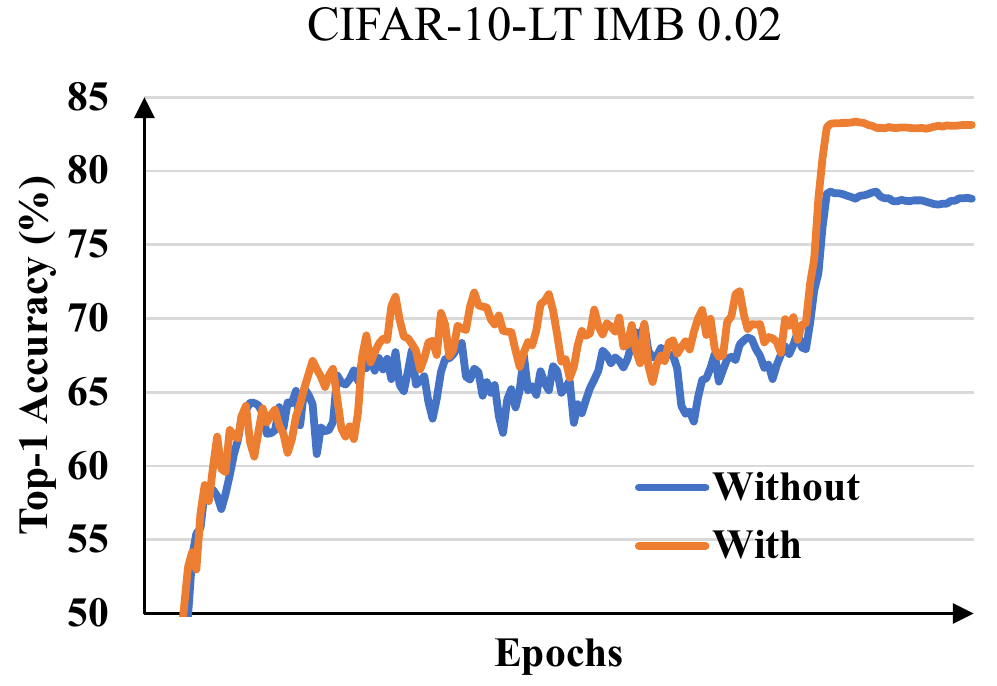}%
		}
		\hspace{0.05in}
		\subfloat[Importance of parameter specialization on CIFAR-100-LT.]{
			\includegraphics[width=0.23\textwidth]{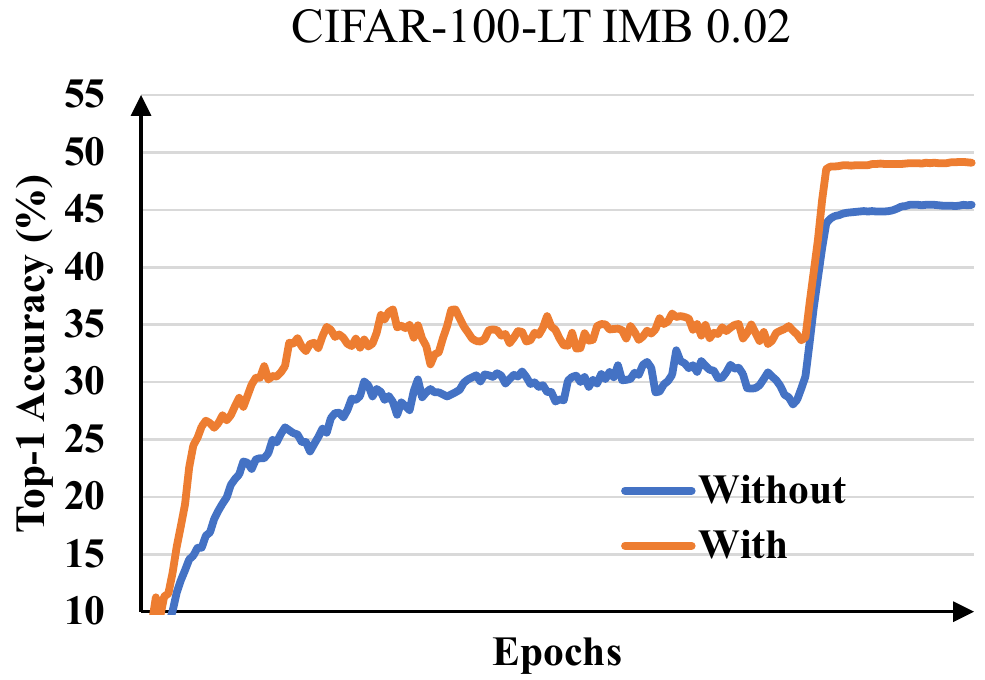} %
		}
	    \hspace{0.05in}
		\subfloat[Residual fusion module analysis on CIFAR-10-LT.]{
			\includegraphics[width=0.23\textwidth]{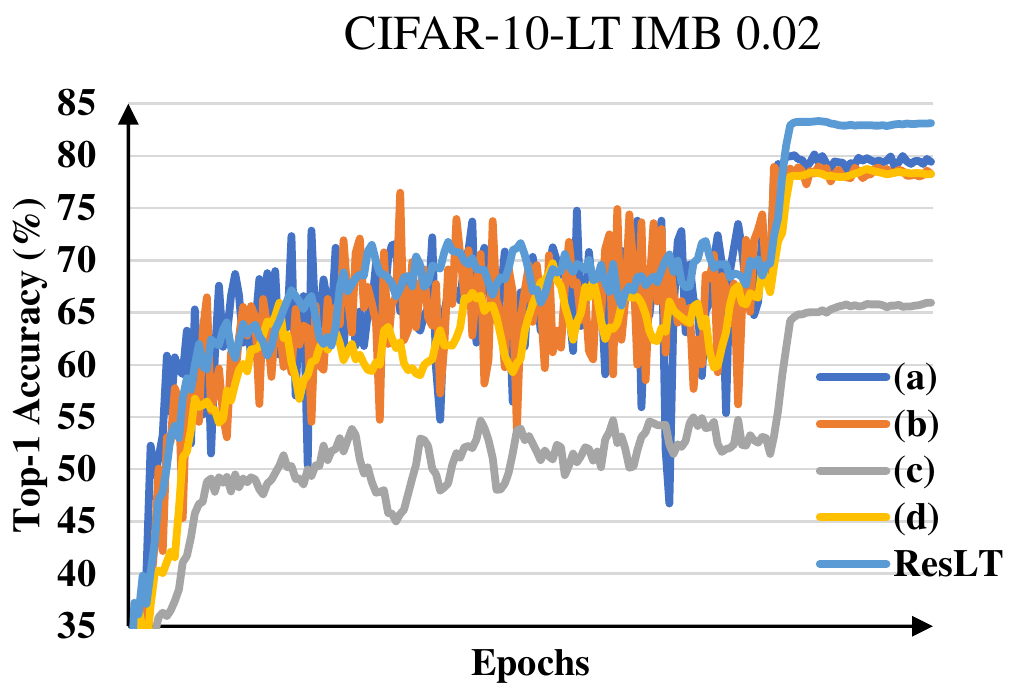}%
		}
		\hspace{0.05in}
		\subfloat[Residual fusion module analysis on CIFAR-100-LT.]{
			\includegraphics[width=0.23\textwidth]{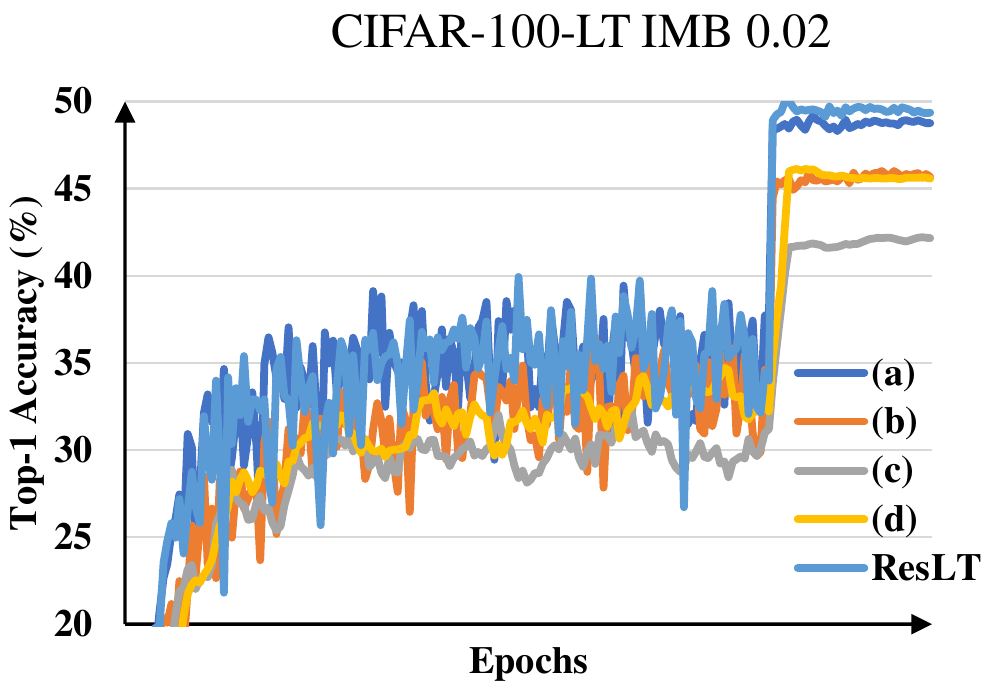} %
		}
		\caption{Ablation studies for the importance of parameter specialization mechanism with (a), (b) and the effectiveness of residual fusion mechanism with (c), (d).}
		\label{fig:ablation_study_reslt}
		\vspace{-0.1in}
	\end{figure*}

	\subsubsection{Importance of Parameter Specialization}
	\label{sec:parameter_specialization}
	Though the three sub-branches are implemented by only a 1x1 grouped convolution with group number 3, we show that it is important to reserve specialized parameters for head, medium and tail classes. We compare strategies with and without the parameter specialization mechanism on CIFAR-10-LT and CIFAR-100-LT with the imbalance factor 0.02. For the setting without parameter specialization, as shown in Fig.~\ref{fig:ablation_study_residual}(e), we let outputs of the three branches be the same while keeping the same loss functions in Eqs. \eqref{eq:res_fusion_loss1}, \eqref{eq:res_fusion_loss2} and \eqref{eq:res_fusion_loss3} in implementation. As shown in Fig.~\ref{fig:ablation_study_reslt}(a) and Fig.~\ref{fig:ablation_study_reslt}(b), the model with the parameter specialization mechanism yields much higher accuracy than the model without it, which manifests the vast importance of our parameter specialization mechanism.
	
	\subsubsection{Analysis of Residual Fusion Module}
	\label{sec:residual_fusion_module}
	To further study the proposed residual fusion mechanism, we conduct ablation study to understand the role of each component in our residual fusion module by comparing its variants on CIFAR-10-LT and CIFAR-100-LT with the imbalance factor 0.02:
	\begin{itemize}
		\item Possible variant (a), (b), (c): as shown in Fig.~\ref{fig:ablation_study_residual}(a), (b), (c), we explore the necessary of the residual mechanism --- the nested classes assignment for different branches in ResLT. $\mathcal{N}_{h}$, $\mathcal{N}_{m}$, $\mathcal{N}_{t}$ and $\mathcal{N}_{h+m+t}$ are optimized to classify images of head, medium, tail and all classes respectively. 
		\item Possible variant (d): as shown in Fig.~\ref{fig:ablation_study_residual}(d), we remove the additive shortcuts from ResLT. During inference, we take the output of main branch $\mathcal{N}_{h+m+t}$ as the final results. The branches $\mathcal{N}_{m+t}$ and $\mathcal{N}_{t}$ only play the role of regularization during training.
	\end{itemize}
	As shown in Fig.~\ref{fig:ablation_study_reslt}(c) and Fig.~\ref{fig:ablation_study_reslt}(d), it is obvious that both the variants (b) and (c) yield lower performance than our full residual fusion module. Variant (a) can enjoy better performance than (b) and (c) but is inferior to ResLT. This phenomenon demonstrates that the residual relationship (nested class assignments) among branch networks and the additive shortcuts are two core factors that make it feasible to solve long-tailed recognition in parameter space. Moreover, it shows the fundamental difference between our proposed residual learning mechanism and previous work \cite{he2016deep}.
	
	\vspace{+0.1in}
	\noindent {\bf Comparison with Re-weighting}
	Our method may reminiscent attentive readers of re-weighting methods considering that the loss with respect to tail class images is calculated by three branches while the loss for head class images is calculated by only $\mathcal{N}_{h+m+t}$ branch. In fact, the comparison between ResLT and the structure of Fig. 2(e) in Sec. \ref{sec:residual_fusion_module} shows the importance of the parameter specialization mechanism. At the same time, it also implies that our high performance doesn't benefit from the naive re-weighting scheme. Specifically, for the structure of Fig. 2(e), the loss for tail class images is also calculated by three branches. However, there is a huge performance gap when compared with the structure of our residual fusion module, {\textit i.e.}, ResLT, as indicated by Fig.~\ref{fig:ablation_study_reslt}(a) and Fig.~\ref{fig:ablation_study_reslt}(b). 
	
	Here we further clarify some fundamental differences between our method and re-weighting. We explicitly preserve specific capacity for the head, medium, tail classes. Besides, re-weighting independently pre-defines scalar factors as the class weights simply according to the number of images in the class while our method adaptively enhances classification results of tail classes with the learned residual branches of $\mathcal{N}_{m+t}$ and $\mathcal{N}_{t}$. Further, our method captures the important relationship between different classes within each branch. 
	
	\vspace{+0.1in}
	\noindent {\bf Discussion on ResLT and RIDE \cite{wang2021longtailed}}
	Training with long-tailed data, deep models can easily overfit to tail classes and thus lead to high variance of model predictions. The motivation for RIDE proposed by Wang et al. is just to reduce the predictions variance by an improved model ensemble strategy. 
	It constructs multiple experts and a distribution-aware diversity loss is deployed to promote high diversity among these experts. 
	In contrast, different from re-weighting rebalancing from loss space and re-sampling rebalancing from input space, ResLT conducts rebalance in the aspect of parameter space by allocating individual parameters for tail classes and train the model with the designed residual mechanism.
	
	Reducing the variance of model predictions is a more general idea. A simple model ensemble strategy can even reduce the predictions variance of a model trained with balanced datasets, {\textit e.g.}, full ImageNet, and thus further improve the performance. Differently, ResLT is just designed for long-tailed data and it should be compatible with techniques that aim to reduce the variance of model predictions. This analysis is coherent with our experimental results shown in Table~\ref{tab:ride_reslt}. Further, the multiple experts structure in RIDE makes it hard to reuse public pre-trained models, {\textit e.g.}, ImageNet pre-trained models that are usually adopted for Places-LT in the literature. Instead, ResLT is more compatible with previous work in this aspect.
	
	\begin{table}	
		\centering
		\caption{Top-1 accuracy (\%) on iNaturalist 2018. Rows with \dag~denote results directly copied from \cite{kang2019decoupling, zhou2019bbn}. $\star$ denotes model is trained with 360 epochs data. "\S" denotes the model is trained with mixup and initialized with ImageNet pre-trained model by torchvision. Knowledge distillation is not applied to all methods for fair comparison.}
		\label{tab:inat_main}
		\begin{tabular}{|c|c|c|}
			\hline
			Method      &One-stage &ResNet-50 \\
			\hline
			\hline
			CB-Focal \cite{cb-focal} \dag         &\ding{52} &61.1 \\
			LDAM \cite{ldam} \dag                 &\ding{52} &64.6 \\
			BBN \cite{zhou2019bbn} $\star$        &\ding{52} &69.6 \\
			$\tau$-normalized \cite{kang2019decoupling}     &\ding{52} &69.3 \\
			Causal Norm \cite{tang2020long}                 &\ding{52} &64.0 \\
			\hline
			\hline
			LDAM+DRW \cite{ldam} \dag                       &\ding{56} &68.0 \\
			cRT \cite{kang2019decoupling}                   &\ding{56} &67.6 \\
			LWS \cite{kang2019decoupling}                   &\ding{56} &69.5 \\
			LWS \cite{kang2019decoupling} \S                &\ding{56} &71.4 \\
			\hline
			\hline
			ResLT             &\ding{52} &70.2 \\
			ResLT $\star$     &\ding{52} &70.5 \\
			ResLT \S          &\ding{52} &72.3 \\
			\hline
		\end{tabular}
	\end{table}
	
	\begin{figure*}[htp]
		\begin{center}
			\includegraphics[width=0.99\textwidth]{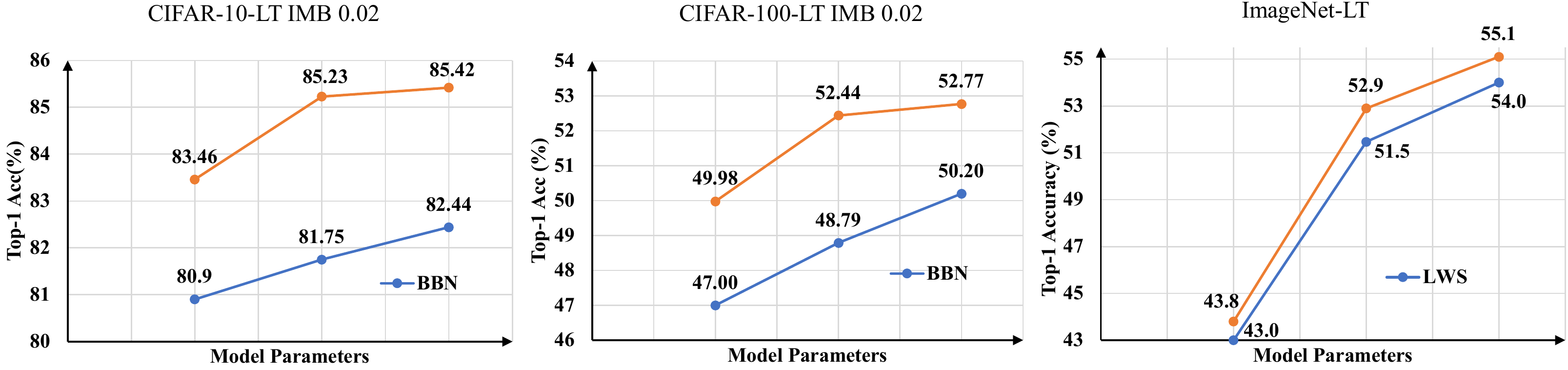}%
		\end{center}
		\vspace{-0.1in}
		\caption{Effects of model size on long-tailed recognition. ResNet-32(1x), ResNet-32(2x), ResNet-32(3x) are used for CIFAR-LT while ResNet-10, ResNeXt-50, ResNeXt-101 are applied for ImageNet-LT.}
		\label{fig:ablation_study_modelsize}
	\end{figure*}
	
	\begin{figure*}[htp]
		\begin{center}
			\includegraphics[width=0.99\textwidth]{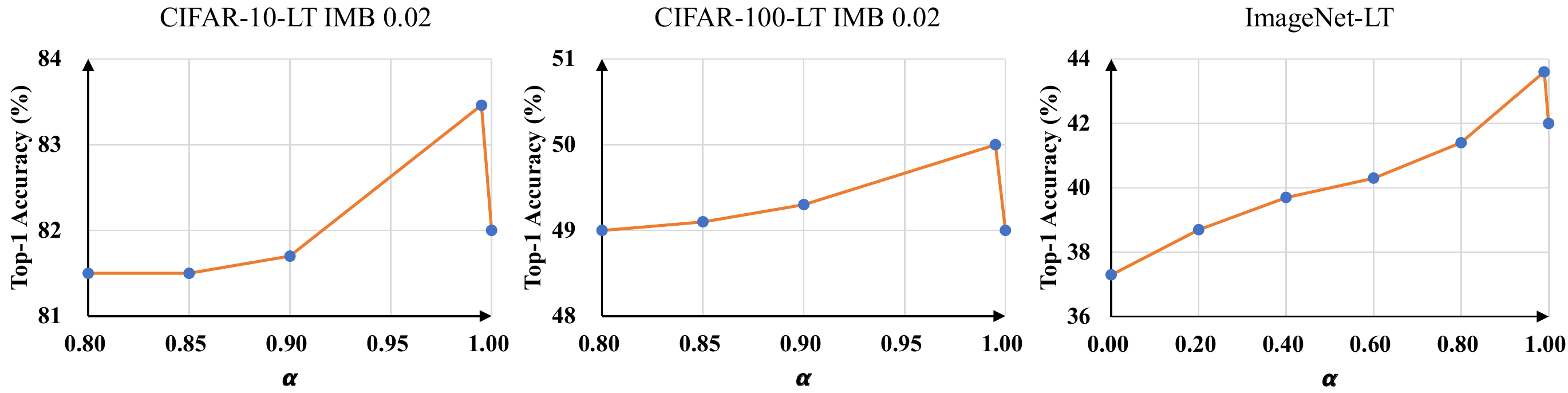}%
		\end{center}
		\vspace{-0.1in}
		\caption{We conduct ablation studies for choices of $\alpha$ on CIFAR-LT and ImageNet-LT. For CIFAR-LT, ResNet-32 is used, while we apply ResNet-10 on ImageNet-LT.}
		\label{fig:ablation_study_alpha}
	\end{figure*}
	
	\begin{table}[htp]
	    \centering
		\caption{Top-1 accuracy (\%) on ImageNet-LT for ResNeXt-50. \dag~denotes results directly copied from \cite{Liu_2019_CVPR}. "\S" denotes the model is trained with strong data augmentation (mixup and autoaugment). Knowledge distillation is not applied to all methods for fair comparison.}
		\label{tab:imagenet_main}
		\resizebox{.95\linewidth}{!}
		{
			\begin{tabular}{|c|c|c|c|c|c|}
				\hline
				Method  & One-stage &Many &Medium &Few &All \\
				\hline
				\hline
				Baseline(CE)                                      &\ding{52} &67.6 &42.8 &13.4 &47.6 \\
				Mixup \cite{mixup}                                &\ding{52} &67.3 &41.0 &11.1 &46.3 \\ 
				Lifted Loss \cite{DBLP:conf/cvpr/SongXJS16} \dag  &\ding{52} &41.1 &35.4 &24.0 &35.2 \\
				Focal Loss \cite{focalloss} \dag                  &\ding{52} &41.1 &34.8 &22.4 &34.6 \\
				Range Loss \cite{zhang2017range} \dag             &\ding{52} &41.1 &35.4 &23.2 &35.1 \\ 
				$\tau$-normalized \cite{kang2019decoupling}       &\ding{52} &60.9 &49.2 &36.8 &51.5\\
				Causal Norm \cite{tang2020long}                   &\ding{52} &64.9 &48.0 &28.9 &51.9 \\
				\hline
				\hline
				cRT \cite{kang2019decoupling}                &\ding{56} &63.7 &47.6 &28.3 &50.7\\
				LWS \cite{kang2019decoupling}                &\ding{56} &63.3 &48.4 &32.0 &51.5\\
				\hline
				\hline
				ResLT               &\ding{52}               &63.0 &50.5 &35.5 &52.9 \\
				ResLT \S            &\ding{52}               &63.6 &55.7 &38.9 &56.1 \\
				\hline
			\end{tabular}
		}
	\end{table}
	
	\begin{table}[htp]
	    \centering
		\caption{Top-1 accuracy (\%) on Places-LT for ResNet-152. \dag~denotes results directly copied from \cite{kang2019decoupling}. "\S" denotes pre-trained model on ImageNet is trained with strong data augmentation (cutmix) provided by \cite{yun2019cutmix}.}
		\label{tab:places_main}
	    \resizebox{.95\linewidth}{!}
	    {
			\begin{tabular}{|c|c|c|c|c|c|}
				\hline
				Method  & One-stage &Many &Medium &Few &All \\
				\hline
				\hline
				Baseline (CE)                                       &\ding{52} & 45.7 & 27.3 & 8.2 & 30.2 \\
				Lifted Loss \cite{DBLP:conf/cvpr/SongXJS16} \dag    &\ding{52} & 41.1 & 35.4 & 24.0 & 35.2 \\
				Focal Loss \cite{focalloss} \dag                    &\ding{52} & 41.1 & 34.8 & 22.4 & 34.6 \\
				Range Loss \cite{zhang2017range} \dag               &\ding{52} & 41.1 & 35.4 & 23.2 & 35.1 \\
				$\tau$-normalized \cite{kang2019decoupling}         &\ding{52} &37.8 &40.7 &31.8 &37.9 \\
				\hline
				\hline
				OLTR \cite{Liu_2019_CVPR}                     &\ding{56} &44.7 &37.0 &25.3 &35.9 \\
				LFME \cite{xiang2020learning}                 &\ding{56} &39.3 &39.6 &24.2 &36.2 \\
				cRT \cite{kang2019decoupling}                 &\ding{56} &42.0 &37.6 &24.9 &36.7 \\
				LWS \cite{kang2019decoupling}                 &\ding{56} &40.6 &39.1 &28.6 &37.6 \\
				RIDE \cite{wang2021longtailed}                &\ding{56}    &-    &-    &-    &- \\
				\hline
				\hline
				ResLT       &\ding{52} &39.8 &43.6 &31.4 &\textbf{39.8} \\
				ResLT \S    &\ding{52} &40.3 &44.4 &34.7 &\textbf{41.0} \\
				\hline
			\end{tabular}
	    }
	\end{table}
	
	\begin{table*}[t]
		\caption{Top-1 accuracy (\%) of ResNet-32 on long-tailed CIFAR-10 and CIFAR-100 (best results are marked in bold font). \dag~denotes results directly copied from \cite{zhou2019bbn}. "\S" denotes the model is trained with strong data augmentation (mixup and autoaugment). Knowledge distillation is not applied to all methods for fair comparison.}
		\begin{center}
			\begin{tabular}{|c|c|c|c|c|c|c|c|}
				\hline
				Dataset          &One-stage &\multicolumn{3}{c|}{Long-tailed CIFAR-10}  &\multicolumn{3}{c|}{Long-tailed CIFAR-100}   \\ 
				\hline
				Imbalance factor  &-  &0.01 &0.02 &0.1   &0.01 &0.02 &0.1      \\ 
				\hline
				\hline
				Mixup \cite{mixup} \dag                         &\ding{52} &73.06 &77.82 &87.10    &39.54 &44.99 &58.02 \\
				Focal \cite{focalloss} \dag                     &\ding{52} &70.38 &76.72 &86.66    &38.41 &44.32 &55.78 \\
				LDAM \cite{ldam}                                &\ding{52} &73.35 &-     &86.96    &39.60 &-     &56.91 \\ 
				BBN  \cite{zhou2019bbn}                         &\ding{52} &79.82 &82.18 &88.32    &42.56 &47.02 &59.12 \\
				CB-Focal \cite{cb-focal} \dag                   &\ding{52} &74.57 &79.27 &87.10    &39.60 &45.17 &57.99 \\
				Causal Norm \cite{tang2020long}                 &\ding{52} &80.60 &83.60 &88.50    &44.10 &50.30 &59.60  \\
				\hline
				\hline
				CE+DRW \cite{ldam}                       &\ding{56} &76.34 &79.97 &87.56 &41.51 &45.29 &58.12 \\
				CE+DRS \cite{ldam}                       &\ding{56} &75.61 &79.81 &87.38 &41.61 &45.48 &58.11 \\
				LDAM+DRW \cite{ldam}                     &\ding{56} &77.03 &81.03 &88.16 &42.04 &46.62 &58.71 \\	
				ELF(LDAM)+DRW \cite{DBLP:journals/corr/abs-2006-11979}   &\ding{56} &78.10 &82.40 &88.00 &43.10 &47.50 &58.90 \\
				RIDE (3 experts) \cite{wang2021longtailed}               &\ding{56} &- &- &- &48.6 &- &- \\                
				\hline
				\hline
				ResLT             &\ding{52}  &80.44   &83.46 &89.06   &45.34 &49.98 &60.79  \\
				ResLT\S           &\ding{52}  &82.40   &85.17 &89.70   &48.21 &52.71 &62.01  \\
			    ResLT (3 experts) &\ding{52}  &- &- &-                 &\textbf{49.73} &\textbf{54.51}  &\textbf{63.73} \\
				\hline
			\end{tabular}
		\end{center}
		\label{tab:cifar_results}
	\end{table*}

	\section{Experiments}
	\subsection{Datasets and Experimental Setting}
	\vspace{+0.1in}
	\noindent {\bf CIFAR-10-LT and CIFAR-100-LT datasets }
	CIFAR-10 and CIFAR-100 both have 60,000 images -- 50,000 for training and 10,000 for validation with 10 categories and 100 categories respectively. For a fair comparison, we use the long-tailed versions of CIFAR datasets with the same setting as those used in \cite{cao2019learning, zhou2019bbn}. 
	Following \cite{zhou2019bbn}, we conduct experiments with imbalance factors 0.01, 0.02, and 0.1.
	
	\vspace{+0.1in}
	\noindent {\bf ImageNet-LT and Places-LT}
	ImageNet-LT and Places-LT were proposed in \cite{Liu_2019_CVPR} by Liu et.al. ImageNet-LT is a long-tailed version of the large-scale object classification dataset ImageNet \cite{imagenet} by sampling a subset following the Pareto distribution with power value $\alpha$=6. It contains 115.8K images from 1,000 categories, with class cardinality ranging from 5 to 1,280. Places-LT is a long-tailed version of the large-scale scene classification dataset Places \cite{zhou2017places}. It consists of 184.5K images from 365 categories with class cardinality ranging from 5 to 4,980.
	
	\vspace{+0.1in}
	\noindent {\bf iNaturalist 2018}
	The iNaturalist 2018 \cite{van2018inaturalist} is one species classification dataset, which is on a large scale and suffers from extremely imbalanced label distributions. It is composed of 437.5K images from 8,142 categories. In addition to the extreme imbalance, iNaturalist 2018 dataset also confronts the fine-grained problem \cite{wei2019piecewise}.
	
	\vspace{+0.1in}
	\noindent {\bf Evaluation Protocol} Following Liu et al. \cite{Liu_2019_CVPR} and Kang et al. \cite{kang2019decoupling}, on ImageNet-LT, Places-LT, and iNaturalist 2018, we report results on three splits of the set of classes: Many-shot (more than 100 images), Medium-shot (20 $\sim$ 100 images) and Few-shot (less than 20 images). For CIFAR-10-LT and CIFAR-100-LT, we sort all the classes by the number of images in decreasing order. Specifically, we divide CIFAR-10-LT into 3 splits as Many-shot (class 0th $\sim$ class 2th), Medium-shot (class 3th $\sim$ class 5th) and Few-shot (class 6th $\sim$ class 9th). For CIFAR-100-LT, we divided it into 3 splits as Many-shot (class 0th $\sim$ class 34th), Medium-shot (class 35th $\sim$ class 69th) and Few-shot (class 70th $\sim$ class 99th).
	
	\vspace{+0.1in}
	\noindent {\bf Training Details}
	For long-tailed CIFAR datasets, we follow \cite{cao2019learning, zhou2019bbn} to pre-process images. We randomly crop a 32x32 patch from the original image or its horizontal flip with 4 pixels padded on each side and normalize the pixel values into [0,1]. To be consistent with previous setting \cite{cao2019learning, zhou2017places}, we adopt ResNet-32 \cite{he2016deep} as our backbone network for all experiments. SGD optimizer with momentum 0.9 is adopted. We train all models for 200 epochs for our ResLT method.  
	The initial learning rate is set to 0.1 and the first five epochs are trained with the linear warm-up. The learning rate is decayed at the 160 and 180 epochs by 0.1 respectively. The batch size 128 is used through all the experiments.
	
	For a fair comparison, we adopt the same experimental setting as \cite{kang2019decoupling} on ImageNet-LT, Places-LT and iNaturalist 2018. For Places-LT, following previous setting \cite{Liu_2019_CVPR, kang2019decoupling}, we choose ResNet-152 as the backbone network, pre-train it on the full ImageNet-2012 dataset (provided by torchvision), and finely tune it for 30 epochs on Places-LT. On ImageNet-LT, we report results with ResNet-10, ResNeXt-50 and ResNeXt-101. For consistency with previous setting, ResNet-50 and ResNet-152 are used for iNaturalist 2018 and we train 200 epochs following \cite{kang2019decoupling}. For all experiments, we use SGD optimizer with momentum 0.9 on 4 NVIDIA 2080Ti GPUs. For ResNet-10, ResNet-50, and ResNeXt-50, we adopt cosine learning rate schedule gradually decaying from 0.1 to 0. We use image resolution $224\times 224$ and batch size 256. For ResNet-152 and ResNeXt-101, we adopt cosine learning rate schedule gradually decaying from 0.05 to 0, image resolution $224\times 224$ and batch size 128 for the limited GPU memory.

	\subsection{Ablation Study}
	\subsubsection{Grouped convolution or 3 separate convolution}
	We conduct ablation studies with an extra grouped convolution or 3 separate convolutions inserted in the top of the ResNet-32 on CIFAR-10-LT and CIFAR-100-LT. As shown in Fig.~\ref{fig:grouped_separate_reslt_ablation}, our method significantly surpasses these two baselines, which indicates our high performance does not come from the small number of extra parameters. 
	
	Moreover, it is worthy to note that BBN [41] also has extra parameters. Our method significantly outperforms BBN as shown in Fig.~\ref{fig:ablation_study_modelsize}.
	
	\subsubsection{How does the model size affect our method?}
	To go deeper in understanding the effects of model size on long-tailed recognition, we explore various models with different parameters on CIFAR100-LT, CIFAR10-LT with imbalance factor 0.02, and ImageNet-LT datasets. For CIFAR100-LT, CIFAR10-LT, we evaluate ResNet-32(1x), ResNet-32(2x) and ResNet-32(3x). Specifically 1x, 2x, 3x mean the multiplier of channels in each layer. For ImageNet-LT, ResNet-10, ResNeXt-50, ResNeXt-101 are used. As shown in Fig.~\ref{fig:ablation_study_modelsize}, with the increase of model parameters, our method can consistently surpass BBN on CIFAR-LT and LWS on ImageNet-LT, which indicates the effectiveness of the proposed ResLT method.

	\begin{table*}[h]
		\caption{Top-1 accuracy (\%) of Many-shot, Medium-shot and Few-shot on ImageNet-LT with various ResNet backbones.  \dag~denotes results directly copied from their original paper.}
		\label{tab:imagenet_details}
		\begin{center}
			\begin{tabular}{|c|c|c|c|c|c|c|}
				\hline
				Method  & Backbone Model & One-stage &Many-shot &Medium-shot &Few-shot &All \\
				\hline
				\hline
				CE (baseline)    &ResNet-10 & \ding{52}          &59.7 &29.4 &5.7 &37.3 \\
				SEQL\dag         &ResNet-10 & \ding{52}          &-    &-    &-   &36.4 \\
				cRT              &ResNet-10 & \ding{52}          &53.8 &41.3 &25.4 &43.2 \\
				$\tau$-normalize &ResNet-10 & \ding{52}          &50.4 &42.1 &26.7 &42.7 \\
				LWS              &ResNet-10 & \ding{56}          &51.8 &41.6 &27.6 &43.0  \\
				ResLT            &ResNet-10 & \ding{52}          &52.3 &41.6 &29.5 &43.8 \\
				\hline
				CE (baseline)    &ResNeXt-50 & \ding{52}          &66.8 &42.0 &14.0 &47.0 \\
				cRT              &ResNeXt-50 & \ding{56}          &63.7 &47.6 &28.3 &50.6 \\
				$\tau$-normalize &ResNeXt-50 & \ding{52}          &62.3 &48.9 &33.7 &51.5 \\
				LWS              &ResNeXt-50 & \ding{56}          &63.3 &48.4 &32.0 &51.5  \\
				ResLT            &ResNeXt-50 & \ding{52}          &63.0 &50.5 &35.5 &53.0 \\
				\hline
				CE (baseline)    &ResNeXt-101-32x4d & \ding{52}         &69.6 &44.6 &15.6 &49.6 \\
				cRT              &ResNeXt-101-32x4d & \ding{56}         &66.2 &50.4 &30.8 &53.3 \\
				$\tau$-normalize &ResNeXt-101-32x4d & \ding{52}         &65.3 &51.5 &35.2 &53.9 \\
				LWS              &ResNeXt-101-32x4d & \ding{56}         &65.7 &51.4 &34.7 &54.0  \\
				ResLT            &ResNeXt-101-32x4d & \ding{52}         &63.3 &53.3 &40.3 &55.1 \\
				\hline
			\end{tabular}
		\end{center}
	\end{table*}
	
	\subsubsection{Individual branch performance}
	\label{sec:individual_branch_performance}
	Our parameter specialization mechanism reserves individual capacity for the head, medium, and tail classes. Along with the proposed residual fusion module, we make it possible to rebalance directly in the aspect of parameter space. To deeply understand the role of residual branches, we collect performance on Many-shot, Medium-shot, and Few-shot of each branch with experiments on CIFAR-100-LT, CIFAR-10-LT with an imbalance ratio of 0.01, and ImageNet-LT. 
	
\begin{figure*}[!t]
		\centering
		\subfloat[Individual branch performance on CIFAR100-LT with imbalance ratio 0.01.]{
			\includegraphics[width=0.46\textwidth]{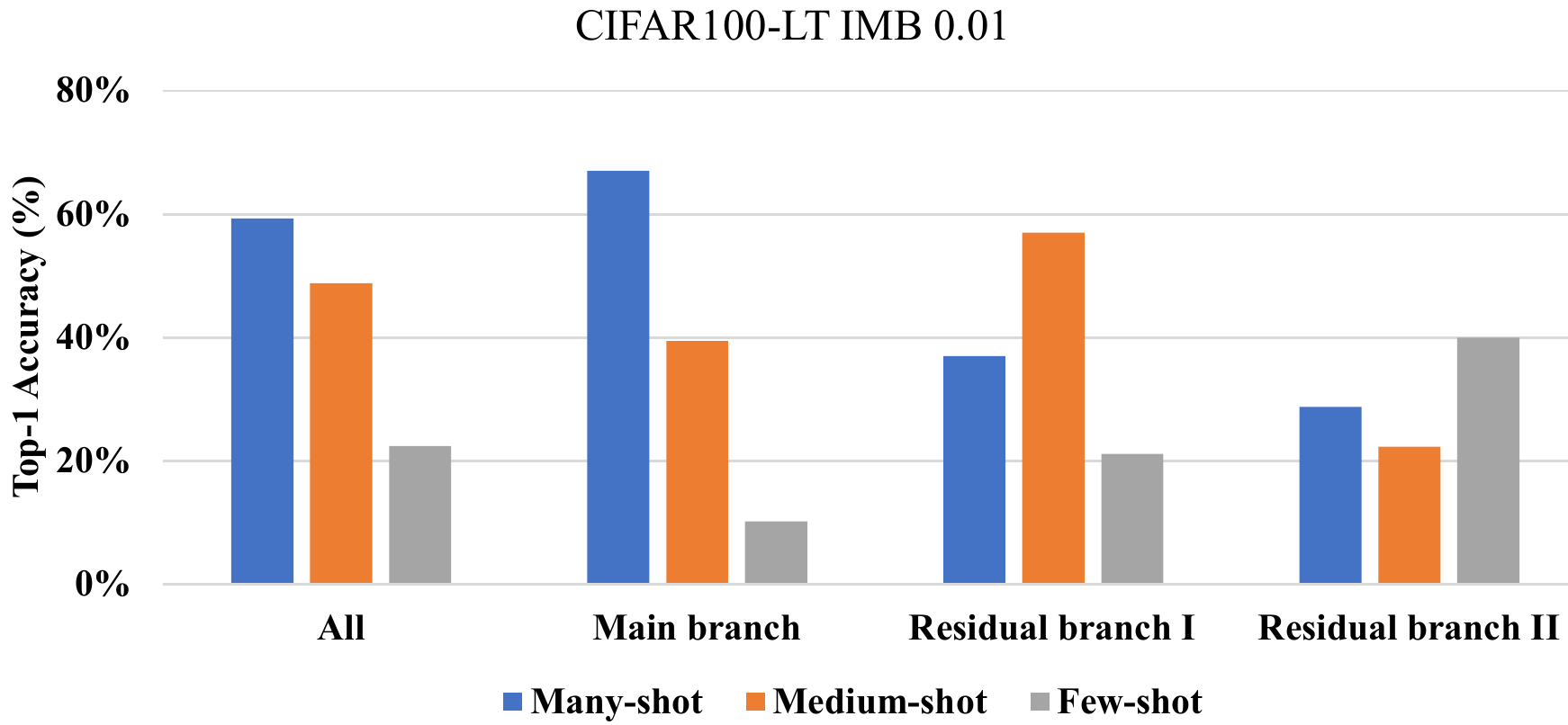}%
		}
		\hspace{0.15in}
		\subfloat[Individual branch performance on CIFAR10-LT with imbalance ratio 0.01.]{
			\includegraphics[width=0.46\textwidth]{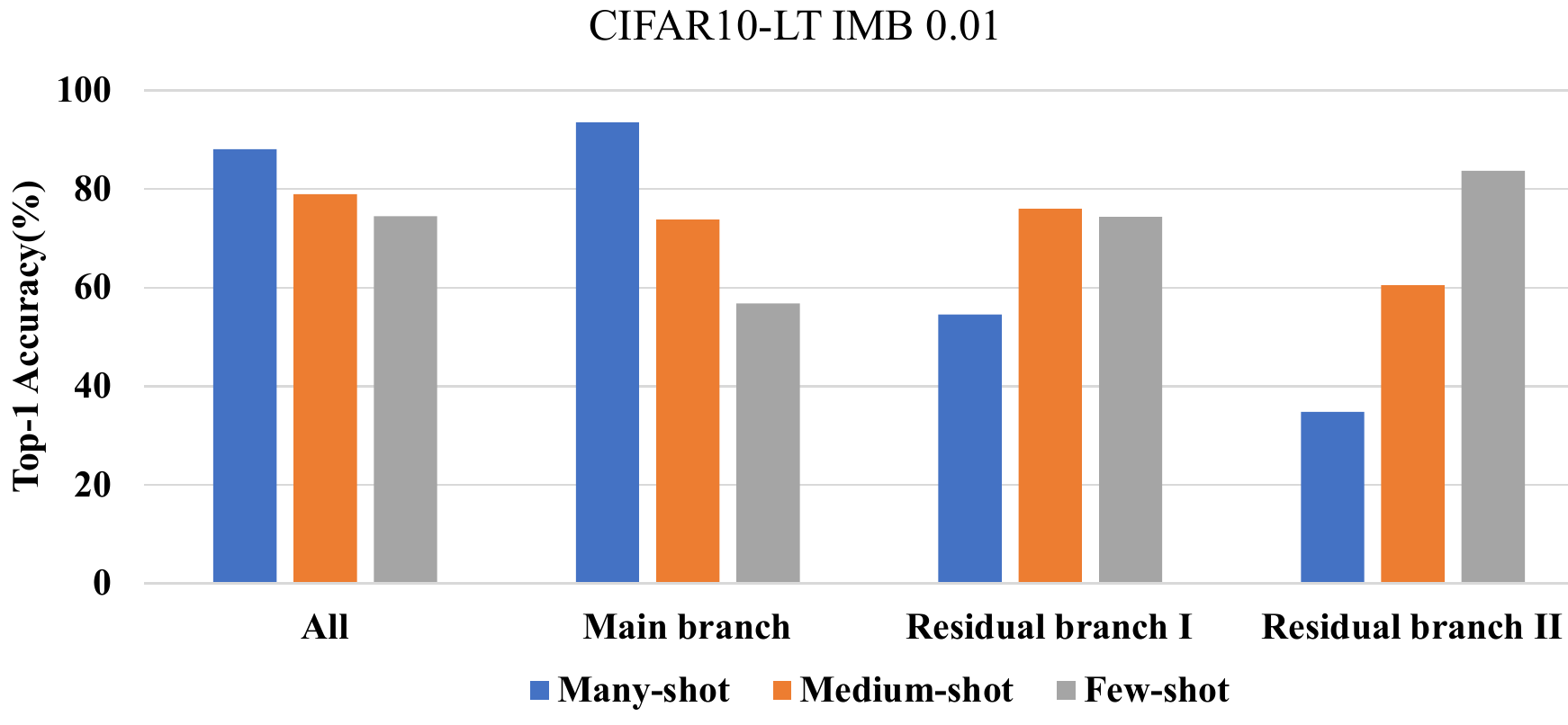} %
		}
		\hspace{0.15in}
		\subfloat[Individual branch performance on ImageNet-LT with ResNet-10.]{
			\includegraphics[width=0.46\textwidth]{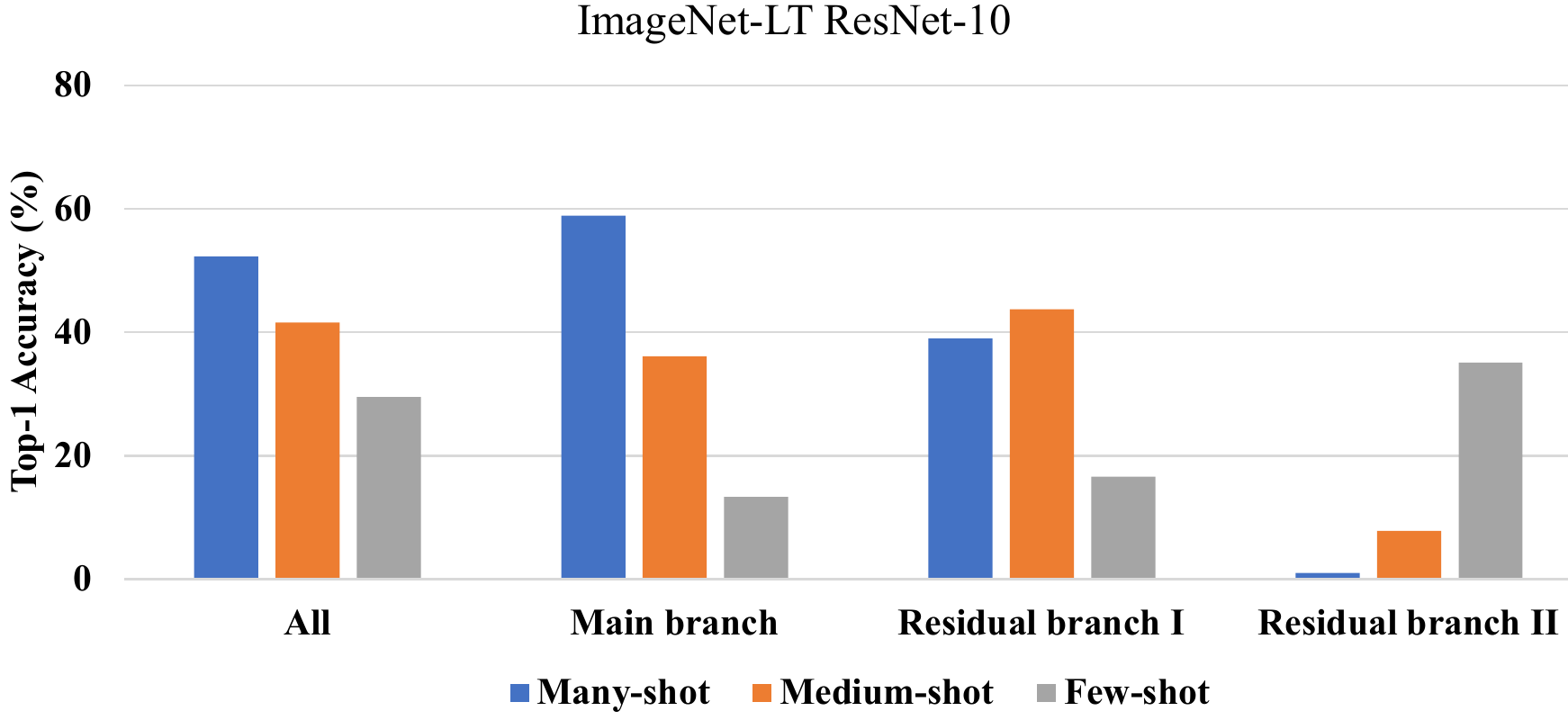} %
		}
		\hspace{0.15in}
		\subfloat[Individual branch performance on ImageNet-LT with ResNeXt-50.]{
			\includegraphics[width=0.46\textwidth]{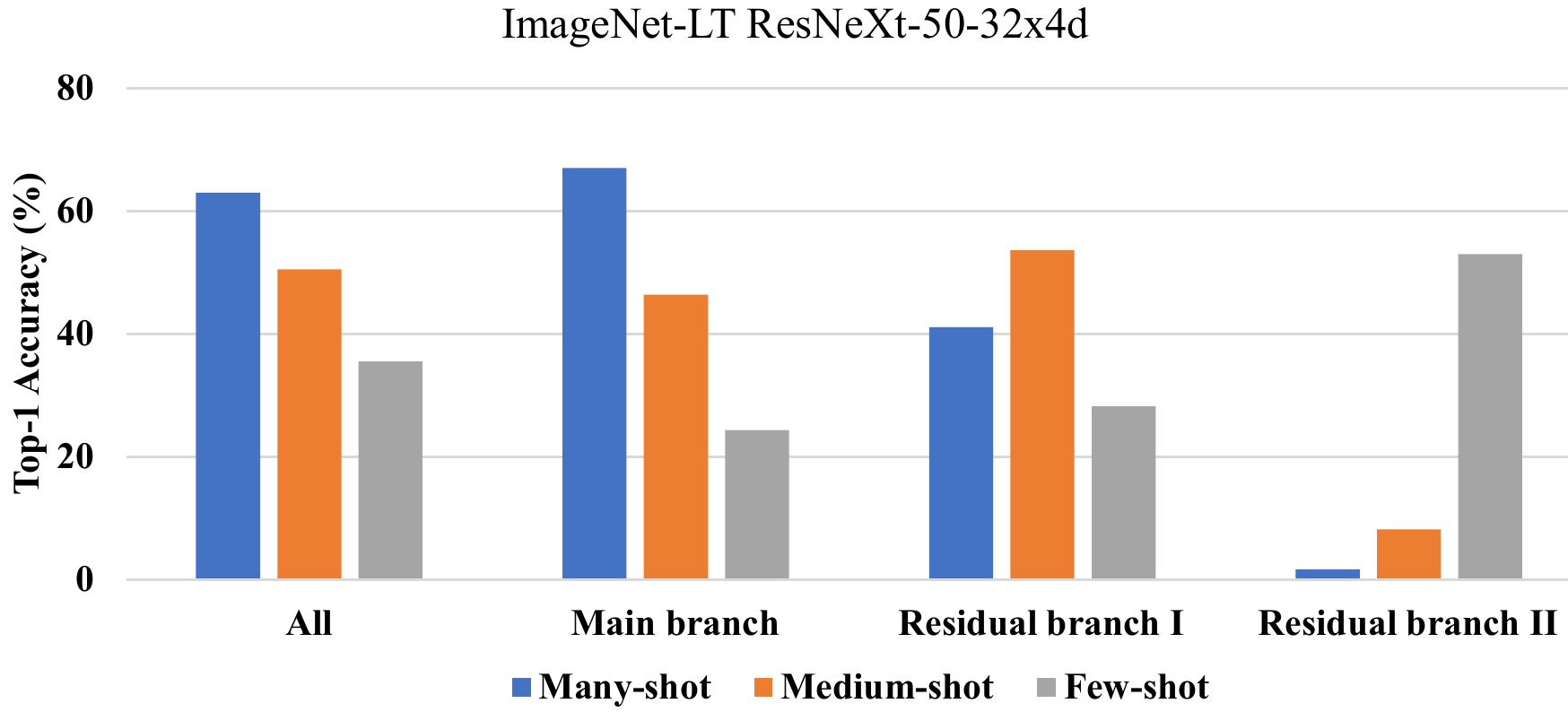}%
		}
		
		\caption{Individual branch performance analysis. ''All'' denotes results are obtained with the aggregated final outputs. Main branch is the $\mathcal{N}_{h+m+t}$. ''Residual branch I'' represents the residual branch $\mathcal{N}_{m+t}$ while ''Residual branch II'' denotes the residual branch $\mathcal{N}_{t}$.}
		\label{fig:ablation_study_individual_branch}
	\end{figure*}

\begin{figure}[htp]
		\centering
		\subfloat[Ablation study for 1-Grouped convolution and 3-Separate convolutions on CIFAR-10-LT.]{
			\includegraphics[width=0.23\textwidth]{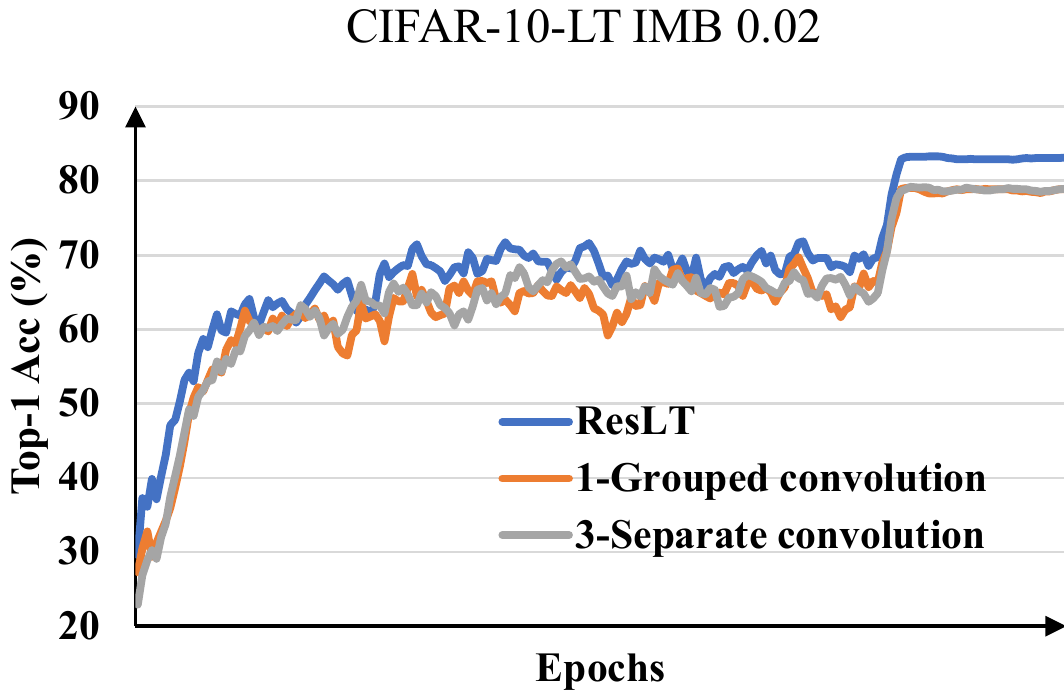}%
		}
		\hspace{0.01in}
		\subfloat[Ablation study for 1-Grouped convolution and 3-Separate convolutions on CIFAR-100-LT.]{
			\includegraphics[width=0.23\textwidth]{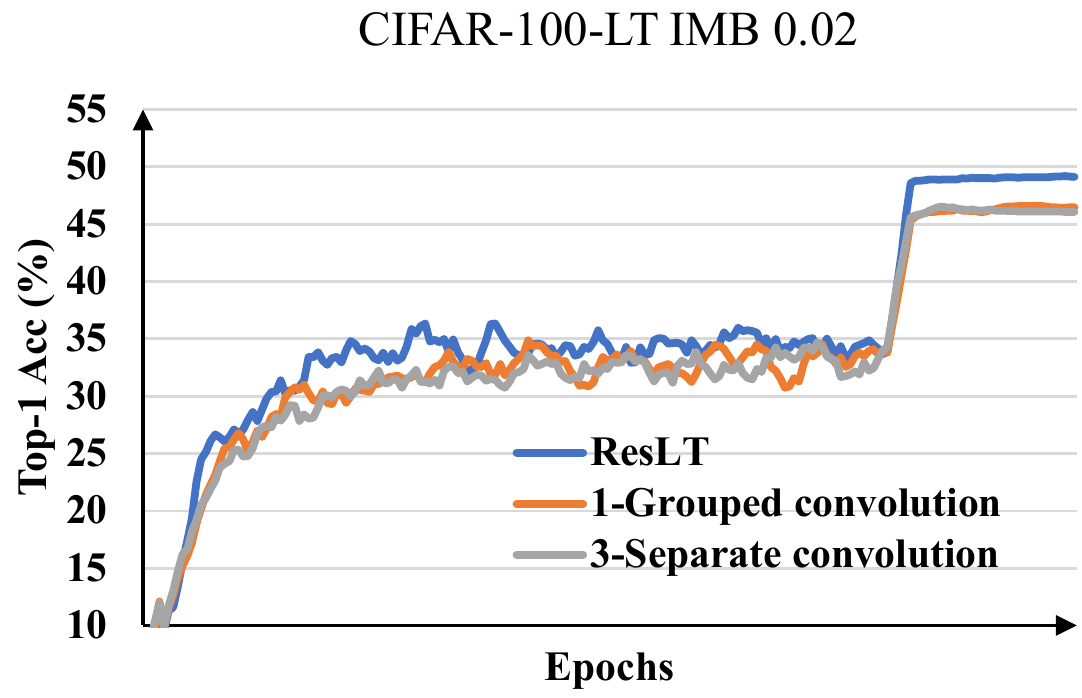} %
		}
		\caption{Ablation study for 1-Grouped convolution and 3-Separate convolutions.}
		\label{fig:grouped_separate_reslt_ablation}
	\end{figure}

	As shown in Figs. \ref{fig:ablation_study_individual_branch}, head classes, medium classes and tail classes respectively dominate branch $\mathcal{N}_{h+m+t}$, $\mathcal{N}_{m+t}$, and $\mathcal{N}_{t}$, realizing the parameter specialization mechanism. Meanwhile, with the two residual branches $\mathcal{N}_{m+t}$ and $\mathcal{N}_{t}$, the final performance on medium classes and tail classes is significantly enhanced compared with the single main branch $\mathcal{N}_{h+m+t}$, testifying the advantages of residual learning mechanism.
	
	\subsubsection{Accuracy on Many-shot, Medium-shot, and Few-shot}
	\label{sec:reslt_better_trade_off}
\begin{figure*}[!t]
		\centering
		\subfloat[Many-shot performance on ImageNet-LT with various ResNet backbones.]{
			\includegraphics[width=0.40\textwidth]{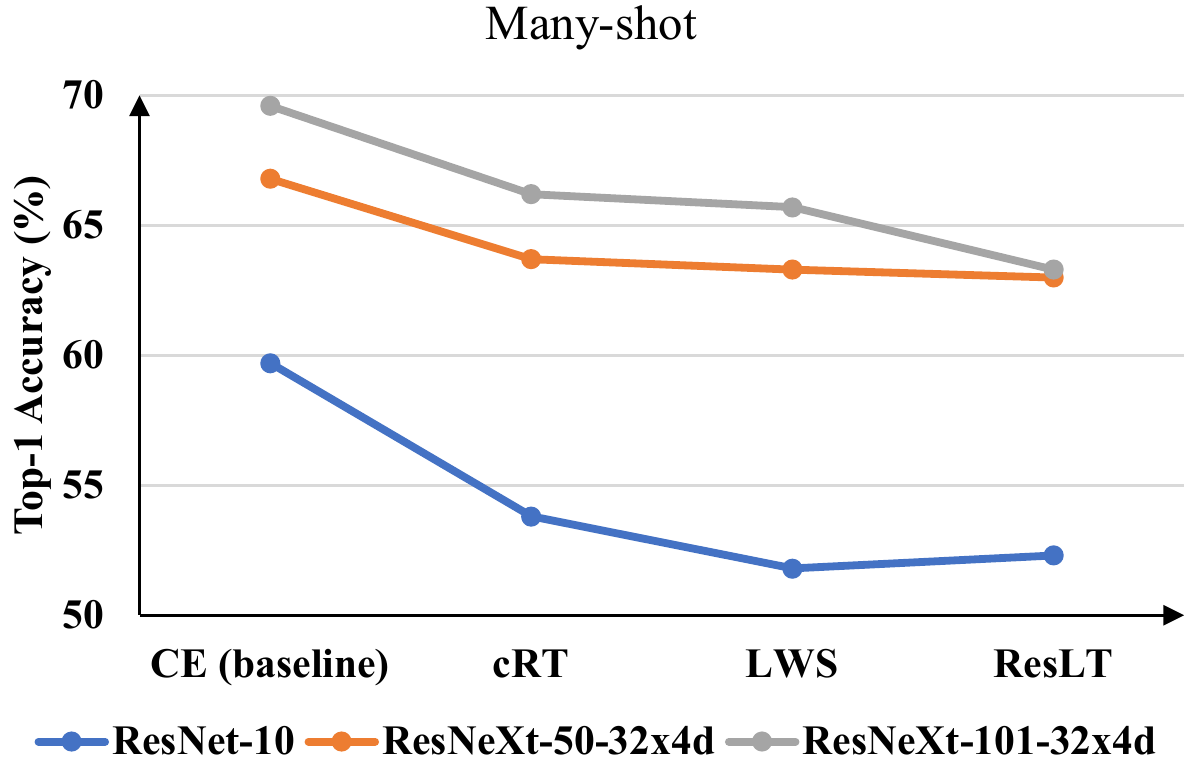}%
		}
		\hspace{0.3in}
		\subfloat[Medium-shot performance on ImageNet-LT with various ResNet backbones.]{
			\includegraphics[width=0.40\textwidth]{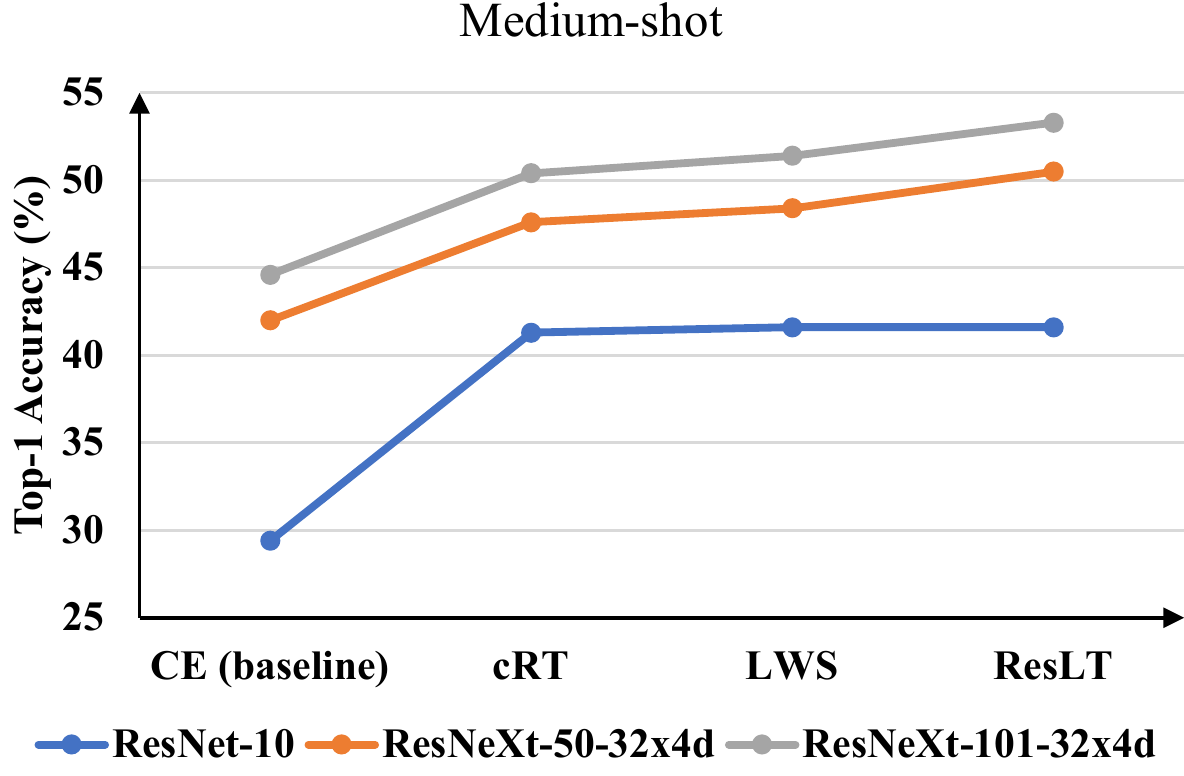} %
		}
		\hspace{0.3in}
		\subfloat[Few-shot performance on ImageNet-LT with various ResNet backbones.]{
			\includegraphics[width=0.40\textwidth]{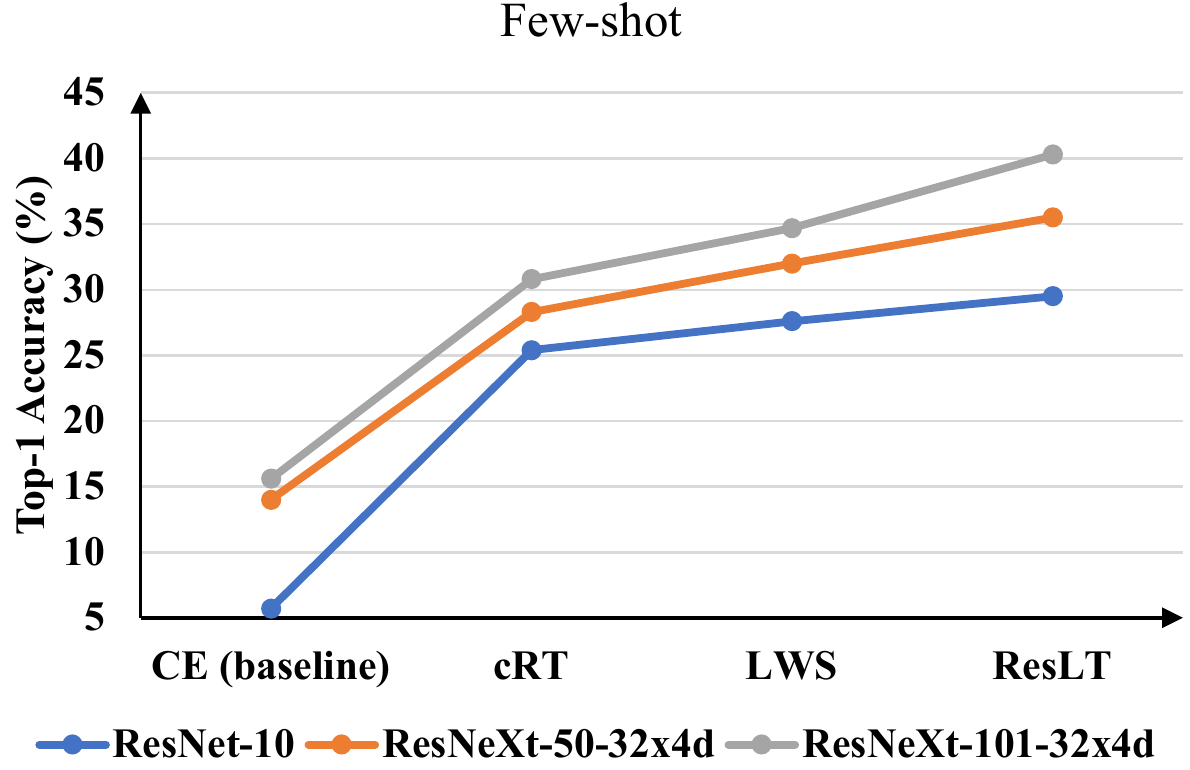} %
		}
		\hspace{0.3in}
		\subfloat[Over-all performance on ImageNet-LT with various ResNet backbones.]{
			\includegraphics[width=0.40\textwidth]{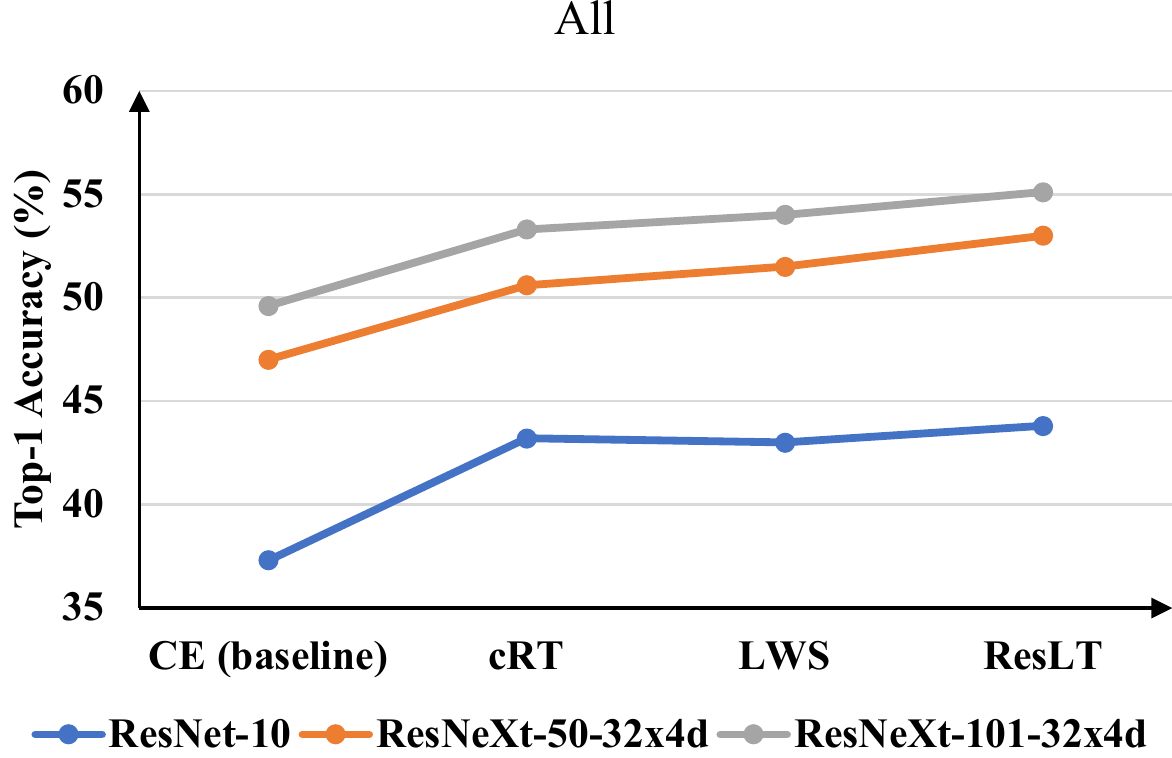}%
		}
		
		\caption{Many-shot, Medium-shot, and Few-shot performance analysis on ImageNet-LT with various ResNet backbones.}
		\label{fig:ablation_study_many_medium_few_shot}
	\end{figure*}

	For ImageNet-LT, the detailed numbers of performance on Many-shot, Medium-shot and Few-shot are summarized in Table~\ref{tab:imagenet_details} while for iNaturalist 2018, the detailed numbers are put in Table~\ref{tab:inat_details}. To highlight the advantages of ResLT, we plot the comparisons on ImageNet-LT with previous methods in Fig~\ref{fig:ablation_study_many_medium_few_shot}. We observe that ResLT usually can achieve higher performance on Medium-shot and Few-shot classes than previous methods, {\it e.g}, LWS, cRT. And the overall improvements just come from small performance degradation on Many-shot classes and such strong performance on Medium-shot and Few-shot classes, which demonstrates ResLT does a better trade-off among Many-shot, Medium-shot, and Few-shot classes. 
	
	\begin{table*}[htp]	
		\centering
		\caption{Top-1 accuracy with 180/200 training epochs and inference speed on ImageNet-LT and iNaturalist 2018. Inference time is calculated with a batch of 64 images on Nvidia GeForce 2080Ti GPU, CUDA11.6, Pytorch1.5, Python3.6.}
		\label{tab:inat_main}
		\begin{tabular}{|c|c|c|c|c|c|c|}
			\hline
			Method  &Epochs   &Speed (ms) &Many-shot &Medium-shot &Few-shot &All\\
			\hline
			\hline
			\multicolumn{7}{|c|}{With RIDEResNeXt-50 on ImageNet-LT} \\
			\hline
			RIDE (2 experts) \cite{wang2021longtailed}     &100 &13.8 &67.3 &50.7 &27.2 &53.9\\
			RIDE (3 experts) \cite{wang2021longtailed}     &100 &17.0 &68.6 &52.2 &27.8 &55.1\\
			RIDE (2 experts) \cite{wang2021longtailed}     &180 &13.8 &66.9 &51.2 &31.2 &54.5\\
			RIDE (3 experts) \cite{wang2021longtailed}     &180 &17.0 &68.4 &52.9 &32.2 &56.0\\
			ResLT (2 experts)     &180 &13.9 &63.3  &55.4  &44.3 &56.4\\
    		ResLT (3 experts)     &180 &17.2 &64.0  &56.6  &44.8 &\textbf{57.6}\\
			\hline
			\hline
			\multicolumn{7}{|c|}{With RIDEResNet-50 on iNaturalist 2018} \\
			\hline
			RIDE (2 experts) \cite{wang2021longtailed}     &100 &12.4 &56.5 &69.9 &70.9 &69.0\\
			RIDE (3 experts) \cite{wang2021longtailed}     &100 &15.9 &62.0 &71.6 &72.0 &70.8\\
			RIDE (2 experts) \cite{wang2021longtailed}     &200 &12.4 &62.5 &71.0 &70.9 &70.1\\
			RIDE (3 experts) \cite{wang2021longtailed}     &200 &15.9 &68.3 &72.6 &71.8 &71.7\\
			ResLT (2 experts)  &200 &12.6 &71.5 &72.0 &72.4 &72.1\\
		    ResLT (3 experts)  &200 &16.2 &73.0 &72.6 &73.1 &\textbf{72.9}\\
			\hline
		\end{tabular}
		\label{tab:ride_reslt}
	\end{table*}
	
	\begin{table*}[htp]
		\caption{Top-1 accuracy (\%) of Many-shot, Medium-shot and Few-shot on iNaturalist 2018 with ResNet-50 backbone.}
		\label{tab:inat_details}
		\begin{center}
			\begin{tabular}{|c|c|c|c|c|c|c|}
				\hline
				Method  & Backbone Model & One-stage &Many-shot &Medium-shot &Few-shot &All \\
				\hline
				\hline
				CE (baseline)    &ResNet-50 & \ding{52}          &75.7 &66.9 &61.7 &65.8 \\
				cRT              &ResNet-50 & \ding{56}          &73.2 &68.8 &66.1 &68.2 \\
				$\tau$-normalize &ResNet-50 & \ding{52}          &71.1 &68.9 &69.3 &69.3 \\
				LWS              &ResNet-50 & \ding{56}          &71.0 &69.8 &68.8 &69.5  \\
				ResLT            &ResNet-50 & \ding{52}          &68.5 &69.9 &70.4 &70.2 \\
				\hline
			\end{tabular}
		\end{center}
	\end{table*}

	\subsubsection{ How does the hyper-parameter $\alpha$ affect our method?}
	\label{sec:alpha}
	We conduct ablation study on choosing a proper $\alpha$ on CIFAR-10-LT, CIFAR-100-LT with imbalance factor 0.02, and ImageNet-LT datasets. For CIFAR-10-LT and CIFAR-100-LT, $\alpha$ = 0.995 is adopted, while we use 0.99 and 0.90 for ImageNet-LT and iNaturalist 2018 respectively in our experiments. As shown in Fig.~\ref{fig:ablation_study_alpha}, even when $\alpha$ = 1 that only with $\mathcal{L}_{branch}$ loss in training, the performance is even much better than baselines in Table~\ref{tab:parameter_sepcialization_results} mentioned in Sec. \ref{sec:parameter_specialization_baselines}, demonstrating the importance of the residual mechanism -- {\it nested class assignments for different branches}.

	\subsubsection{How does the number of groups affect our method?}
	\label{sec:groups}
	Here we have further explored the effects of the different number of groups on CIFAR-10-LT, CIFAR-100-LT with an imbalance factor 0.01. As shown in Table~\ref{tab:ablation_study_number_groups}, when the number of groups is larger than 3, there are no obvious improvements.
	
	\begin{table}[h]
		\caption{ Ablation study for different number of groups. "\#" represents the number of groups. Top-1 accuracy (\%) are reported. We use ResNet-32 as backbone in experiments.}
		\label{tab:ablation_study_number_groups}
		\begin{center}
			\begin{tabular}{|c|c|c|c|c|}
				\hline
				Dataset      &\#2 &\#3 &\#4 &\#5 \\
				\hline
				CIFAR-10-LT  &78.83 &80.44 &80.80 &80.44  \\
				\hline
				CIFAR-100-LT &45.11 &45.34 &45.34 &45.14  \\
				\hline
			\end{tabular}
		\end{center}
	\end{table}
	
	\subsubsection{Feature visualizations}
\begin{figure*}[htp]
		\centering
		\subfloat[CIFAR-10-LT with ResLT]{
			\includegraphics[width=0.28\textwidth]{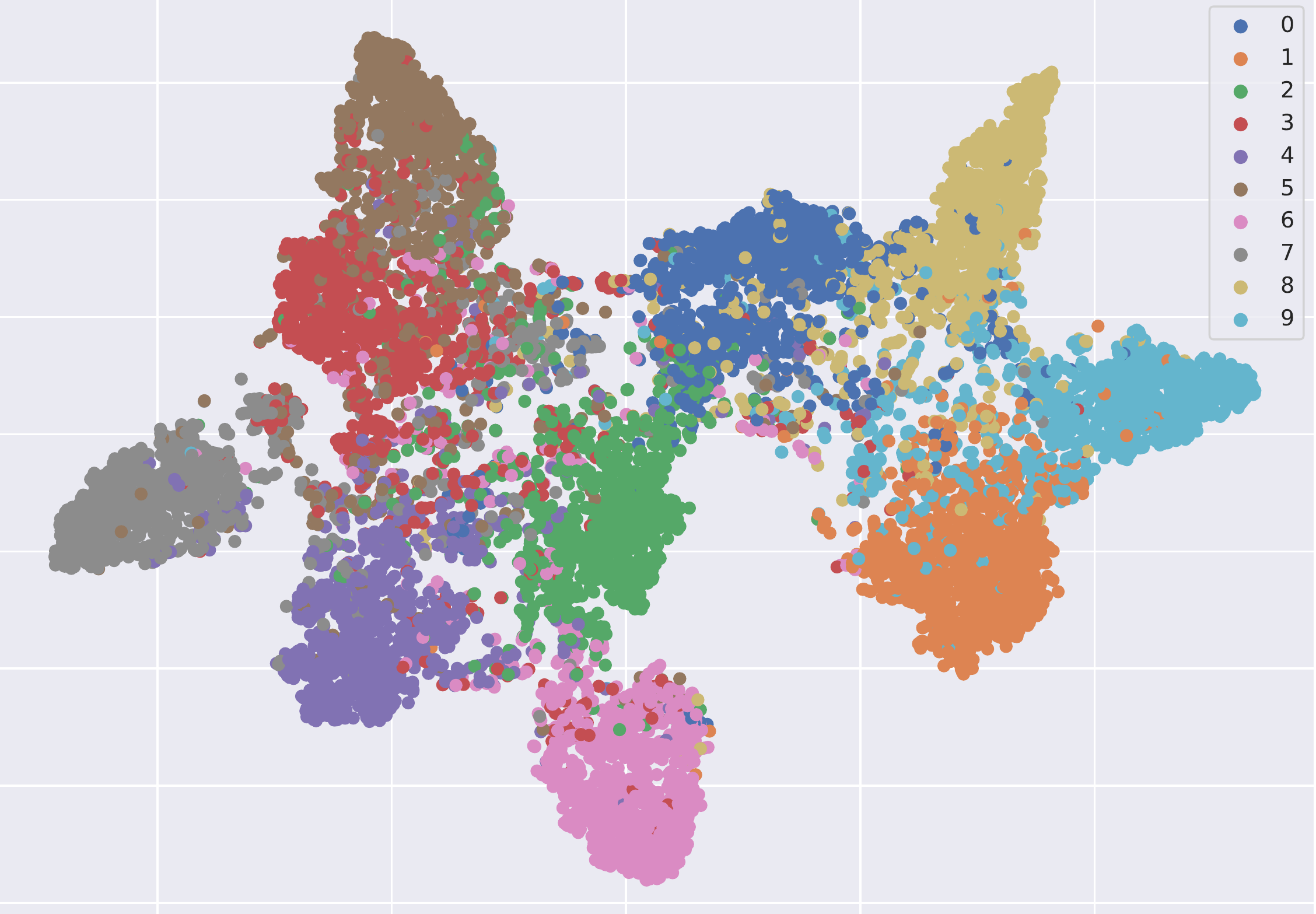}%
		}
		\hspace{0.1in}
		\subfloat[CIFAR-100-LT random selected 10 classes with ResLT ]{
			\includegraphics[width=0.28\textwidth]{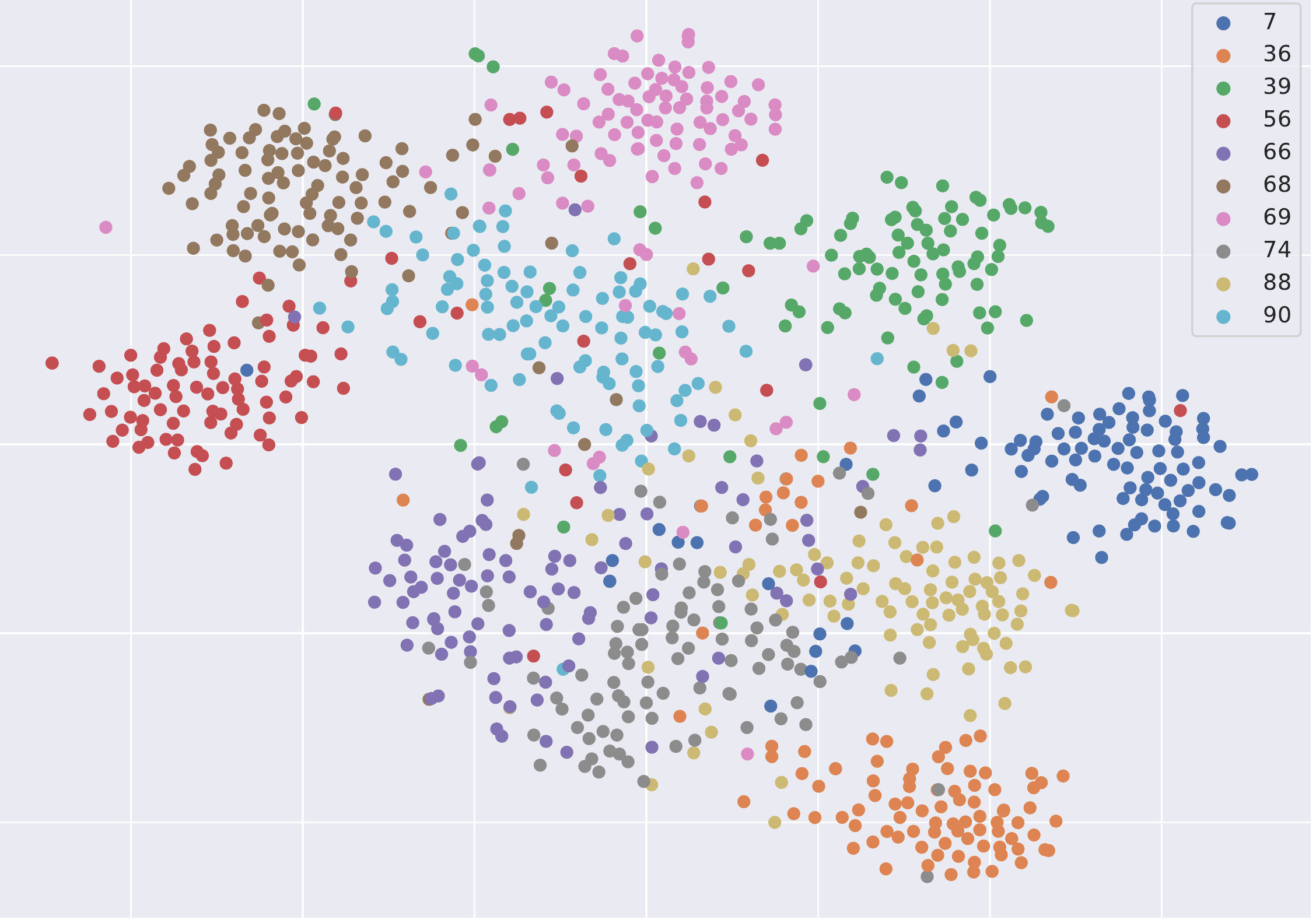} %
		}
		\hspace{0.1in}
		\subfloat[CIFAR-100-LT random selected 20 classes with ResLT ]{
			\includegraphics[width=0.28\textwidth]{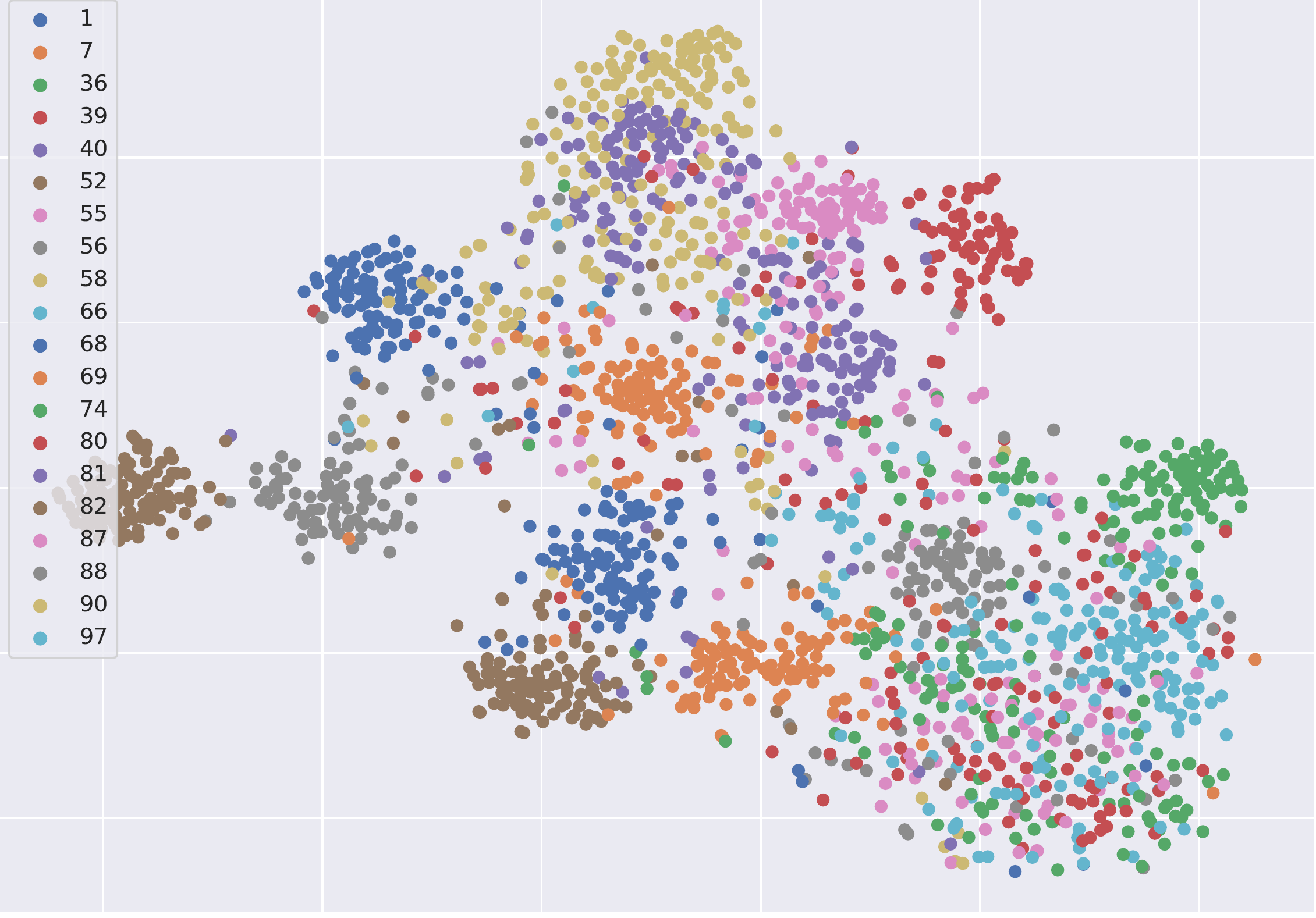} %
		}
		\hspace{0.1in}
		\subfloat[CIFAR-10-LT with cross-entropy]{
			\includegraphics[width=0.28\textwidth]{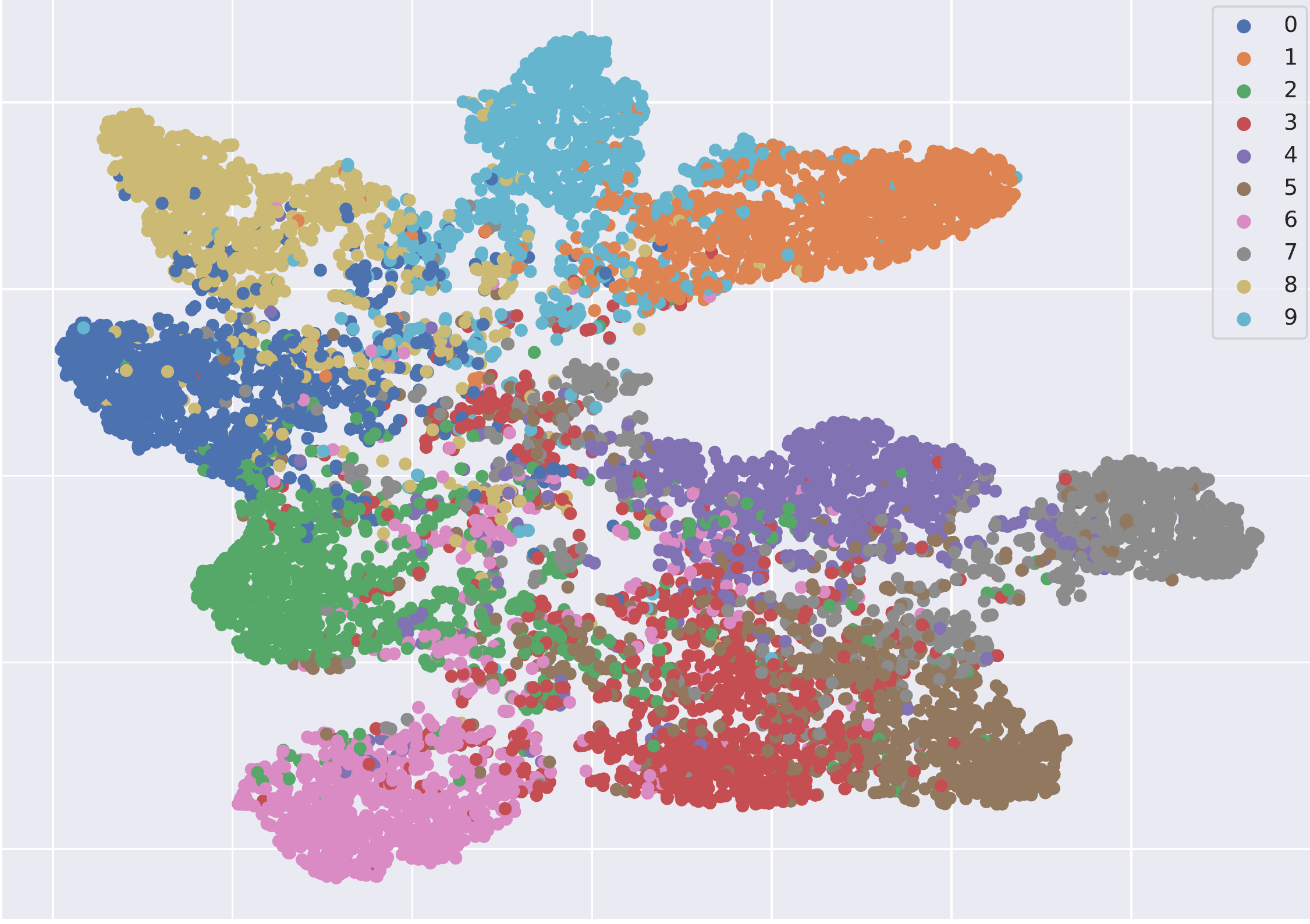}%
		}
		\hspace{0.1in}
		\subfloat[CIFAR-100-LT random selected 10 classes with cross-entropy.]{
			\includegraphics[width=0.28\textwidth]{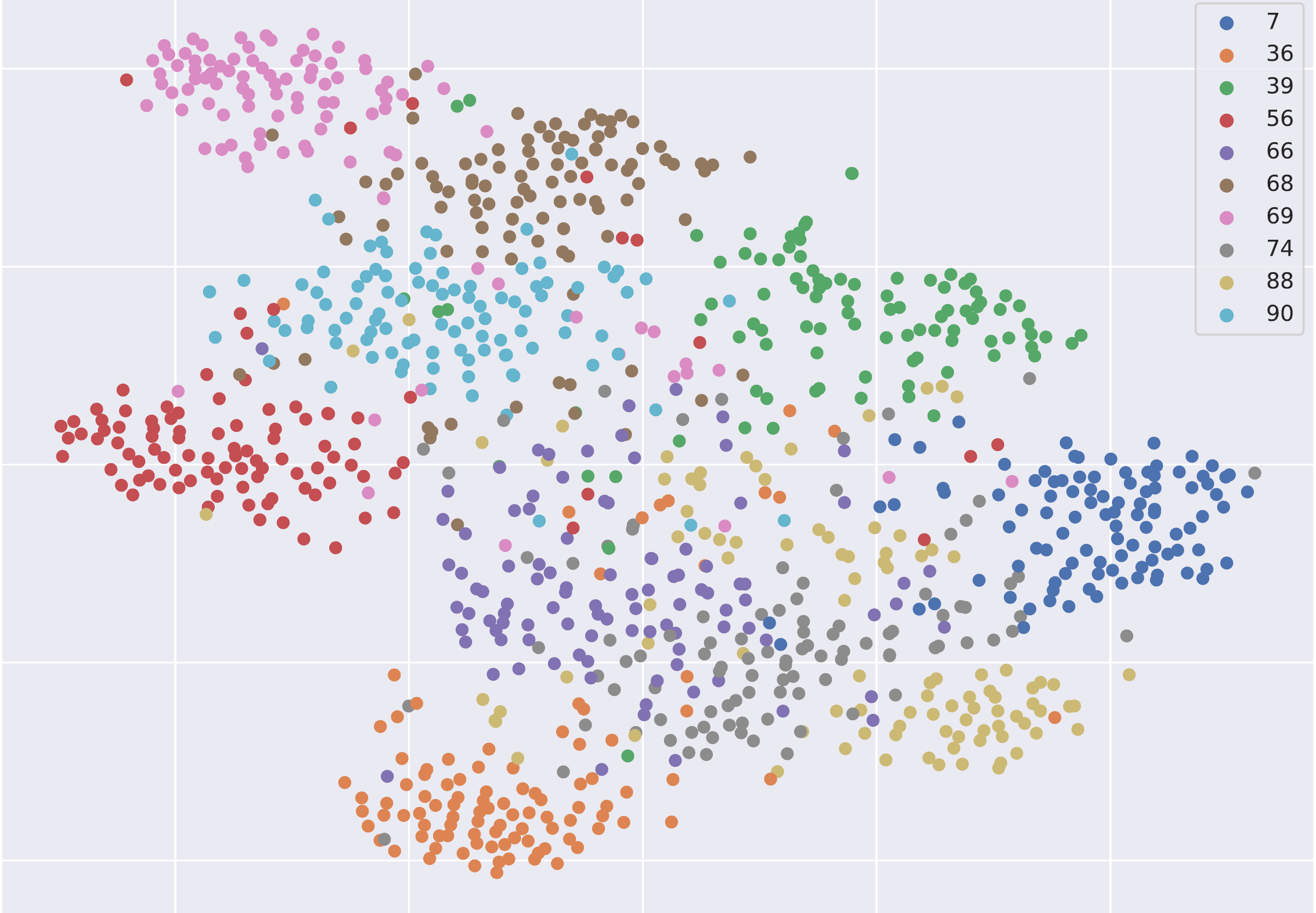}%
		}
		\hspace{0.1in}
		\subfloat[CIFAR-100-LT random selected 20 classes with cross-entropy.]{
			\includegraphics[width=0.28\textwidth]{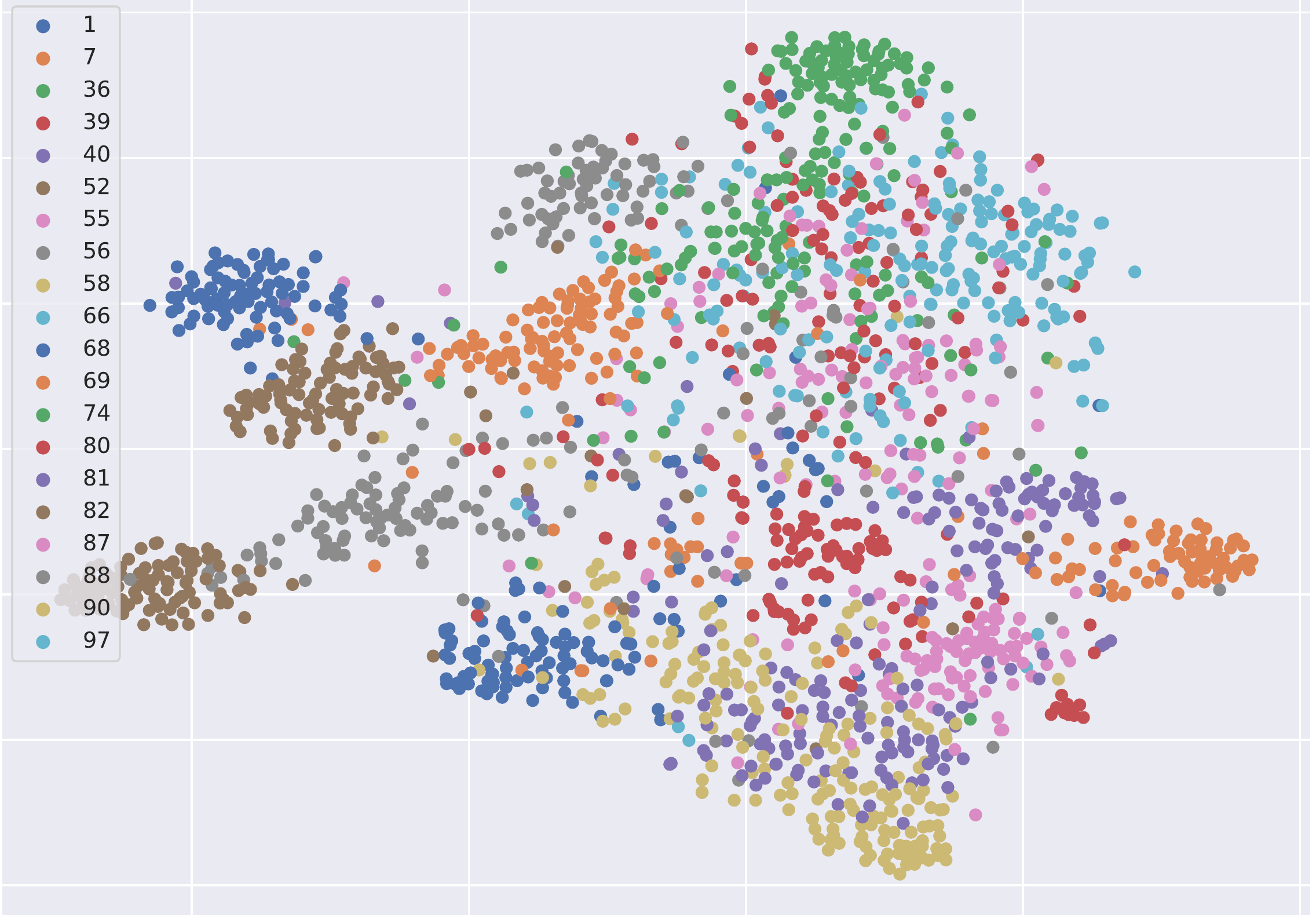}%
		}
		\caption{Visualization comparison between ResLT trained model and cross-entropy trained model. Numbers in all the figures are class indexes.}
		\label{fig:ablation_study_feature_visualization}
	\end{figure*}

	To go deeper into understanding the effects of ResLT method, we visualize representation learned by vanilla cross-entropy and ResLT training method on CIFAR-10-LT and CIFAR-100-LT with an imbalance ratio of 0.01. Specifically, for CIFAR-100-LT, we visualize the randomly selected 10 classes or 20 classes. As shown in Fig~\ref{fig:ablation_study_feature_visualization}, The models trained with ResLT usually can produce more compact features, leading to more obvious separation and better performance.  
	
	\subsubsection{More discussion on ResLT}
	Zhang et al. \cite{zhang2019theoretically} identified the trade-off between model robustness and accuracy. For long-tailed classification, Cao et al. \cite{ldam} derived a margin-based generalization bound for the optimal trade-off between a head class and a tail class margins in the binary classification problem, which also implied that there was a trade-off between head classes accuracy and tail classes accuracy. As discussed in Sec. \ref{sec:reslt_better_trade_off}, we observe ResLT can learn a better trade-off between head classes accuracy and tail classes accuracy with the proposed residual learning mechanism. We leave more theoretical analysis as our future work.
	
	\subsection{Comparisons with Previous Methods}
	\vspace{+0.1in}
	\noindent{\bf Comparison methods} 
	In experiments, we compared with two kinds of methods: one-stage methods (with a consistent training strategy throughout the training) and state-of-the-art two-stage methods (input sampling strategy or loss function may differ in different stages).
	\begin{itemize}
		\item One-stage methods. For one-stage methods, we mainly compare with mixup \cite{mixup}, LDAM \cite{cao2019learning}, BBN \cite{zhou2019bbn} and Causal Norm \cite{tang2020long} methods.
		\item Two-stage methods. For two-stage methods, we mainly compare with the recently proposed decoupling representation and classifier learning \cite{kang2019decoupling} and RIDE \cite{wang2021longtailed}. For fair comparison with RIDE, we report the experimental results without knowledge distillation trick.
	\end{itemize}
	
	\vspace{+0.1in}
	\noindent {\bf Comparison on ImageNet-LT}
	Table~\ref{tab:imagenet_main} shows experimental results on ImageNet-LT. Here, we mainly compare with the recent SOTA method \cite{kang2019decoupling, tang2020long, wang2021longtailed}. 
	We observe that 90 epochs training is not enough to converge for deep models on ImageNet-LT.
	Under longer training with 180 epochs, Decouple method \cite{kang2019decoupling} can be significantly further improved while it seems to be helpless for Causal Norm \cite{tang2020long}. We use this strong setting for comparisons.
	As shown in Table~\ref{tab:imagenet_main}, ResNext-50 trained with our ResLT method in an end-to-end fashion enjoys 1.4\% higher than \cite{kang2019decoupling} and 1.0\% higher than \cite{tang2020long}. Moreover, as shown in Fig.~\ref{fig:ablation_study_modelsize}, we also test our method on ResNet-10 and ResNext-101. Our method consistently surpasses LWS \cite{kang2019decoupling} by 0.8\% $\sim$ 1.4\% across small models to large models.
	
	For fairly comparing with RIDE, we replace ResNeXt-50 with RIDEResNeXt-50 adopted by RIDE. As shown in Table~\ref{tab:ride_reslt}, ResLT model achieves 57.6\% top-1 accuracy with 3 experts in 180 epochs training, largely outperforming RIDE by 1.6\%. With RIDEResNeXt-50 and 3 experts, our model only needs an extra 0.04G FLOPs computational cost at inference time. We also calculate the inference time with an image size 224 $\times$ 224 in Table~\ref{tab:ride_reslt}. Our model costs 17.2ms when fed with a batch of 64 images while RIDE uses a comparable time of 17.0 ms.

	\vspace{+0.1in}
	\noindent {\bf Comparison on Places-LT}
	Table~\ref{tab:places_main} lists experimental results on Places-LT. Following previous setting \cite{Liu_2019_CVPR, kang2019decoupling}, we use ResNet-152 pre-trained on full ImageNet dataset as our backbone (provided by torchvision). As shown in Table~\ref{tab:places_main}, the model trained with our ResLT in an end-to-end manner achieves 39.8\% Top-1 accuracy, which is 1.9\% higher than previous Decoupling methods \cite{kang2019decoupling}.
	
	\vspace{+0.1in}
	\noindent {\bf Comparison on iNaturalist 2018}
	Experimental results on iNaturalist 2018 are summarized in Table~\ref{tab:inat_main}. We mainly compare our method with BBN and Kang et al. \cite{kang2019decoupling} with 200 training epochs. Note that, though Zhou et al. \cite{zhou2019bbn} claimed BBN(2x) trained with 180 epochs, they actually used 360 epochs data due to their dual sampler strategy. Under the same amount of data for training, our method outperforms \cite{zhou2019bbn} by 0.9\%. Compared with \cite{kang2019decoupling}, our ResLT model trained in an end-to-end manner achieves 70.2\% top-1 accuracy, surpassing it by 0.7\%. Further, with larger network ResNet-152, our method achieves 73.2\% top-1 accuracy, surpassing \cite{kang2019decoupling} by 0.7\% under same training setting. When compared with RIDE, ResLT model achieves 72.9\% top-1 accuracy with 3 experts in 200 epochs training schedule, significantly surpassing RIDE by 1.2\% as shown in Table~\ref{tab:ride_reslt}.
	
	\vspace{+0.1in}
	\noindent {\bf Comparison on CIFAR-10-LT and CIFAR-100-LT}
	We conduct extensive experiments on CIFAR-10 and CIFAR-100 with imbalance factors of 0.01, 0.02 and 0.1, same as previous setting \cite{cao2019learning, zhou2019bbn}.
	The experimental results are summarized in Table~\ref{tab:cifar_results}. Our method outperforms BBN by 2.78\%, 2.96\% and 1.76\% on CIFAR-100-LT with imbalance factors 0.01, 0.02, and 0.1 respectively. Compared with Causal Norm by Tang et al., ResLT surpasses it by 1.24\% and 1.19\% separately with imbalance ratio 0.01 and 0.1 on CIFAR-100. With RIDEResNet, ResLT model boosts performance to 49.73\% on CIFAR-100 under a imbalance ratio 0.01, surpassing RIDE.

	\section{Conclusion}
	In this paper, we provide a novel perspective to understand and address the long-tailed recognition problem. The proposed residual fusion mechanism effectively re-balances head and tail classes under this new point of view in parameter space. Extensive experimental results on various representative and challenging benchmarks manifest the effectiveness of our method.
	
	To understand our method thoroughly, we conduct sound ablation studies to show that parameter specialization and residual learning mechanism are two key components to make it work. And we leave more theoretical analysis for ResLT as our future work.

	\ifCLASSOPTIONcaptionsoff
	\newpage
	\fi

	{\small
		\bibliographystyle{IEEEtran}
		\bibliography{egbib}
	}
	
	\begin{IEEEbiography}
		[{\includegraphics[height=1.25in,clip,keepaspectratio]
			{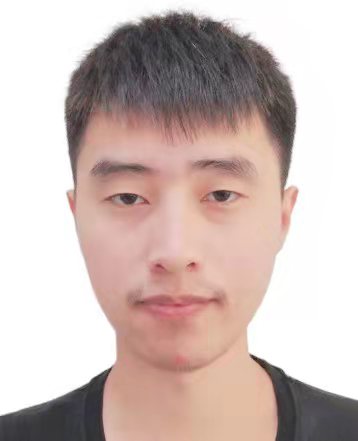}}]
		{Jiequan Cui}
		received the B.Eng. degree (Honors) in Computer Science from the School of Computer Science and Technology, ShanDong University (SDU) in 2018. He is
		currently a Ph.D. Candidate at the Chinese University of Hong Kong (CUHK), under the supervision of Prof. Jiaya Jia. He serves as a reviewer for IJCV, CVPR, ICCV, ECCV, ICLR, NeurIPS. His research interests include model generalization, neural architecture search, adversarial robustness, imbalanced learning, and image segmentation.
	\end{IEEEbiography}
	
	\begin{IEEEbiography}
		[{\includegraphics[width=1in,height=1.25in,clip,keepaspectratio]
			{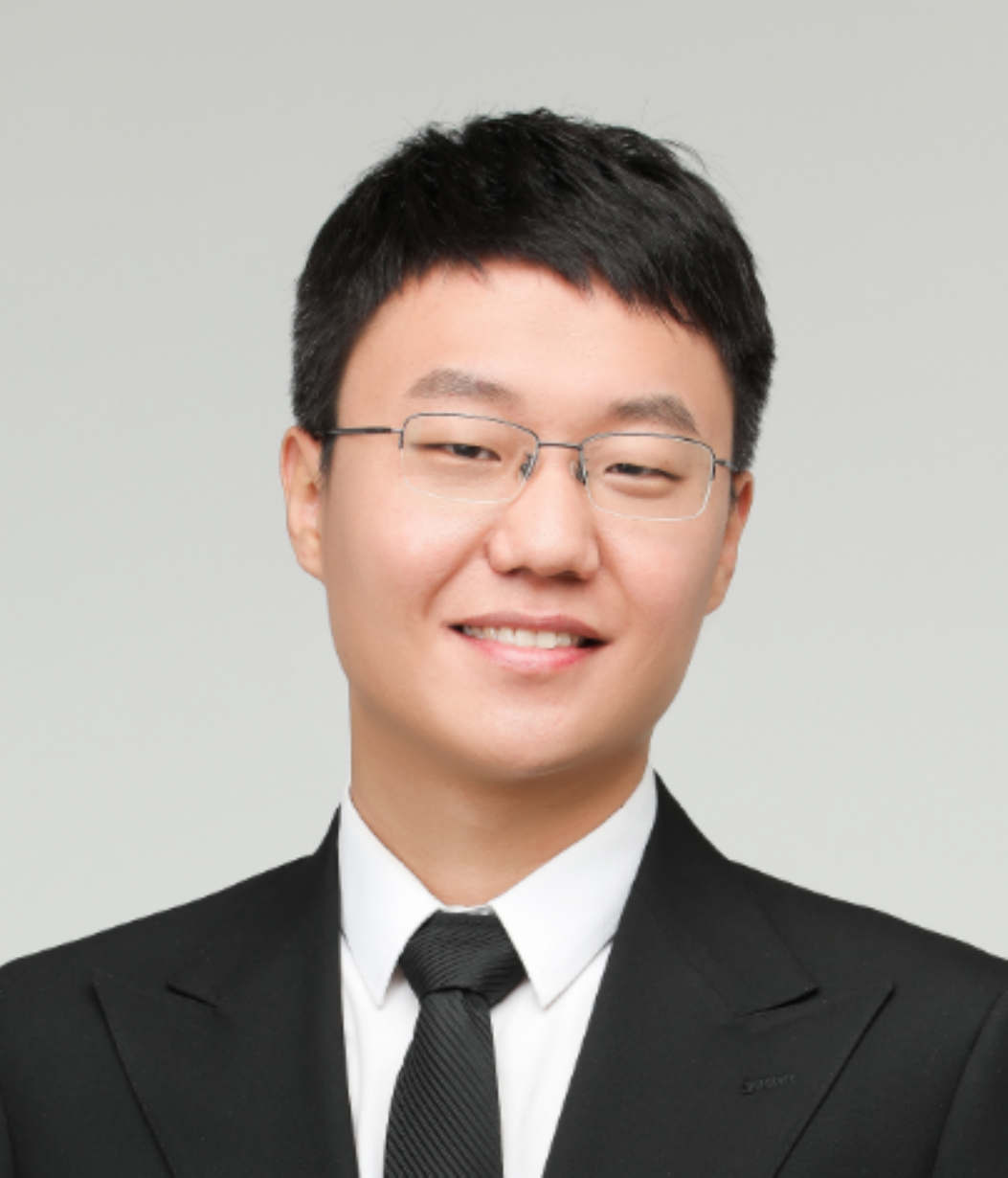}}]
		{Shu Liu}
		now serves as Co-Founder and Technical Head in SmartMore. He received the BS degree from Huazhong University of Science and Technology and the PhD degree from the Chinese University of Hong Kong. He was the winner of 2017 COCO Instance Segmentation Competition and received the Outstanding Reviewer of ICCV in 2019. He continuously served as a reviewer for TPAMI, CVPR, ICCV, NeurIPS, ICLR and etc. His research interests lie in deep learning and computer vision.
	\end{IEEEbiography}	
	
	\begin{IEEEbiography}
		[{\includegraphics[height=1.25in,clip,keepaspectratio]
			{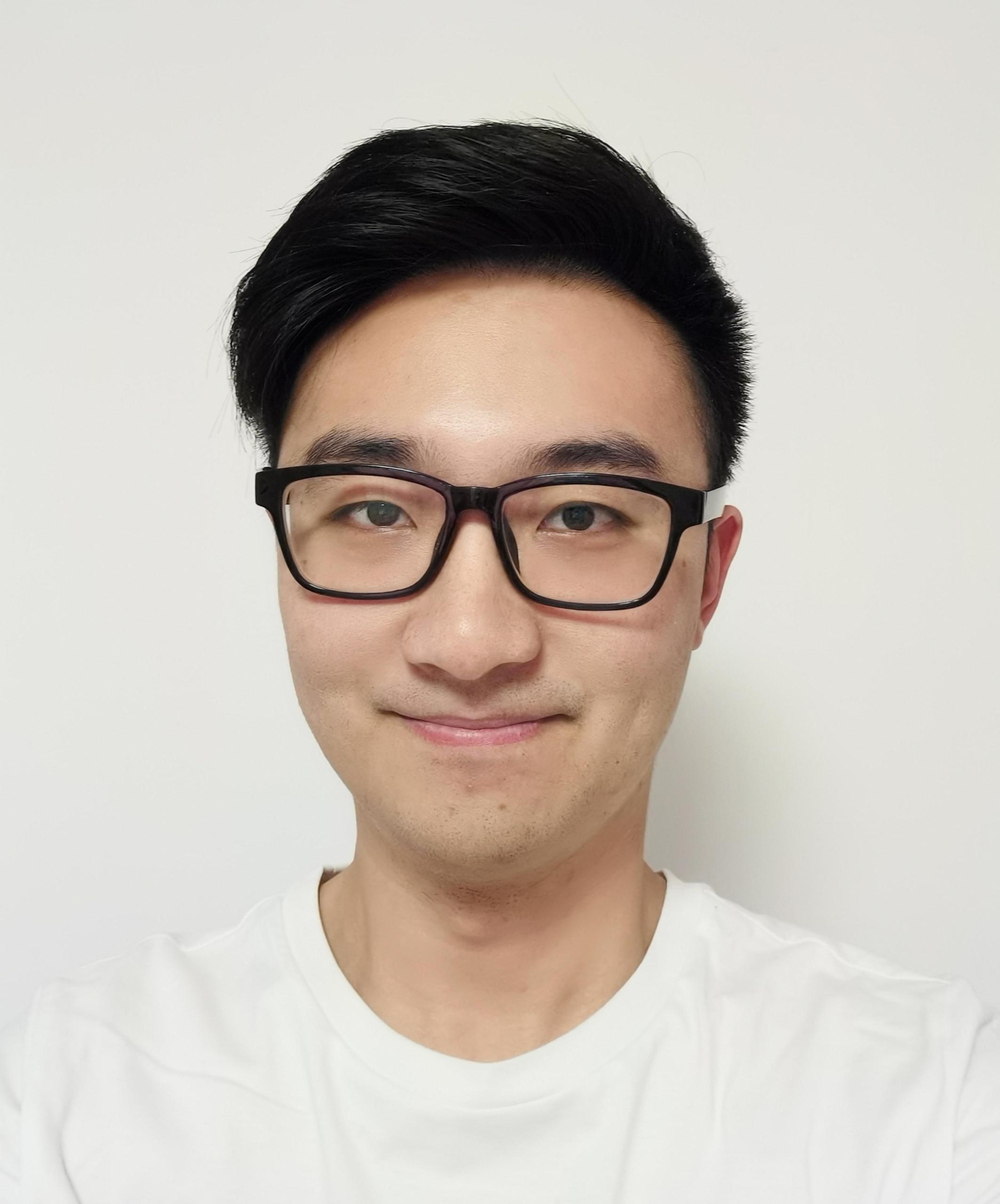}}]
		{Zhuotao Tian}
		received the B.Eng. degree (Honors) in Computer Science from the School of Computer Science and Technology, Harbin Institute of Technology (HIT) in 2018. He is
		currently a 3rd year Ph.D. student at the Chinese University of Hong Kong (CUHK), under the supervision of Prof. Jiaya Jia. He serves as a reviewer for IJCV, CVPR, ICCV, ECCV, AAAI. His research interests include few-shot learning, semi-supervised learning, semantic segmentation and scene text detection.
	\end{IEEEbiography}
	
	\begin{IEEEbiography}
		[{\includegraphics[height=1.25in,clip,keepaspectratio]
			{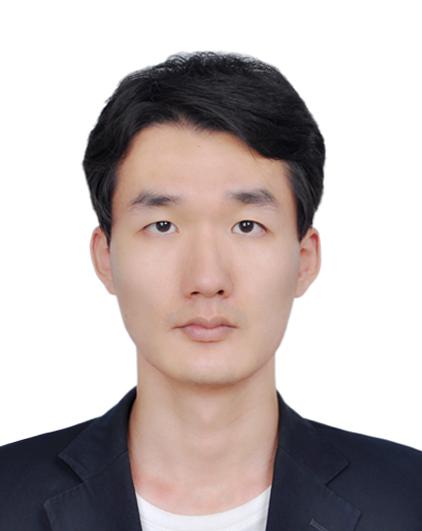}}]
		{Zhisheng Zhong}
		received the B.Eng. Degree in Communication Engineering from Beijing University of Posts and Telecommunications (BUPT) in 2016. He received the Master Degree in Computer Science from Peking University (PKU) in 2019. Now he is a Ph.D. student at the Department of Computer Science Engineering~(CSE), the Chinese University of Hong Kong (CUHK). He serves as a reviewer for NeurIPS, CVPR, ICCV, ICLR and etc. His research interests lie in deep learning and computer vision.
	\end{IEEEbiography}

	\begin{IEEEbiography}
		[{\includegraphics[height=1.25in,clip,keepaspectratio]{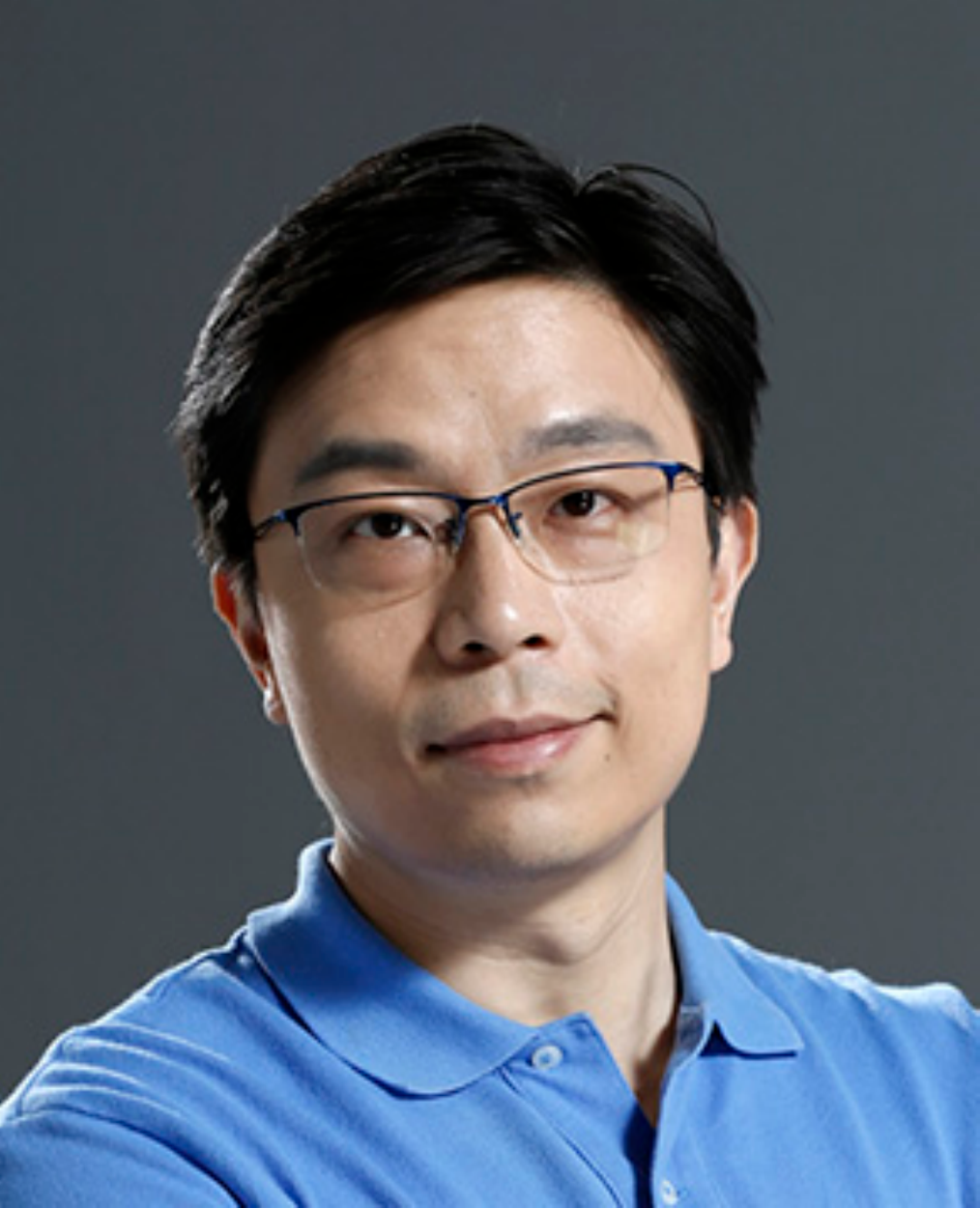}}]
		{Jiaya Jia}
		received the Ph.D.~degree in Computer Science from Hong Kong University of Science and Technology in 2004 and is currently a full professor in Department of Computer Science and Engineering at the Chinese University of Hong Kong (CUHK). He assumes the position of Associate Editor-in-Chief of IEEE Transactions on Pattern Analysis and Machine Intelligence (TPAMI) and is in the editorial board of International Journal of Computer Vision (IJCV). He continuously served as area chairs for ICCV, CVPR, AAAI, ECCV, and several other conferences for the organization. He was on program committees of major conferences in graphics and computational imaging, including ICCP, SIGGRAPH, and SIGGRAPH Asia. He is a Fellow of the IEEE. 
	\end{IEEEbiography}
	
\end{document}